%% file: main.tex
\documentclass{article} % For LaTeX2e
\usepackage{iclr2026_conference,times}

\input{math_commands.tex}

\usepackage{hyperref}
\usepackage{cleveref}
\usepackage{url}
\usepackage{subcaption}
\usepackage{xcolor}
\usepackage{booktabs}
\usepackage{graphicx}
\usepackage{adjustbox}
\usepackage{multirow}
\usepackage{arydshln}
\usepackage{wrapfig} 
\usepackage{array} 
\usepackage[dvipsnames]{xcolor}
\title{Learning Brain Representation with Hierarchical Visual Embeddings}

\author{
Jiawen Zheng\textsuperscript{1}\thanks{Equal technical contribution.}, Haonan Jia\textsuperscript{2}\footnotemark[1], Ming Li\textsuperscript{3}, Yuhui Zheng\textsuperscript{1}, Yufeng Zeng\textsuperscript{1}, Yang Gao\textsuperscript{2}, Chen Liang\textsuperscript{1}\thanks{Corresponding author.} \\
\textsuperscript{1}The Hong Kong University of Science and Technology (Guangzhou)  \\
\textsuperscript{2}Beihang University \quad
\textsuperscript{3}Tsinghua University \\
{\texttt{jzhengbx@connect.hkust-gz.edu.cn}, \texttt{chenliang2@hkust-gz.edu.cn}} \\
}

\iclrfinalcopy % Uncomment for camera-ready version, but NOT for submission.
\begin{document}

\maketitle

\input{sec/0_abstract}

\input{sec/1_introduction}

\input{sec/2_related_work}

\input{sec/3_method}

\input{sec/4_experiment}

\input{sec/5_conclusion}

% \section{General formatting instructions}
% \label{gen_inst}

% The text must be confined within a rectangle 5.5~inches (33~picas) wide and
% 9~inches (54~picas) long. The left margin is 1.5~inch (9~picas).
% Use 10~point type with a vertical spacing of 11~points. Times New Roman is the
% preferred typeface throughout. Paragraphs are separated by 1/2~line space,
% with no indentation.

% Paper title is 17~point, in small caps and left-aligned.
% All pages should start at 1~inch (6~picas) from the top of the page.

% Authors' names are
% set in boldface, and each name is placed above its corresponding
% address. The lead author's name is to be listed first, and
% the co-authors' names are set to follow. Authors sharing the
% same address can be on the same line.

% Please pay special attention to the instructions in section \ref{others}
% regarding figures, tables, acknowledgments, and references.

% There will be a strict upper limit of \textbf{9 pages} for the main text of the initial submission, with unlimited additional pages for citations. This limit will be expanded to \textbf{10 pages} for rebuttal/camera ready.

% Third level headings are in small caps,
% flush left and in point size 10. One line space before the third level
% heading and 1/2~line space after the third level heading.

% \clearpage

\bibliography{iclr2026_conference}
\bibliographystyle{iclr2026_conference}

\clearpage

\appendix
\input{sec/6_appendix}

\end{document}

%% file: math_commands.tex
\usepackage{amsmath,amsfonts,bm}

\def\eqref#1{equation~\ref{#1}}
\def\1{\bm{1}}

\DeclareMathAlphabet{\mathsfit}{\encodingdefault}{\sfdefault}{m}{sl}
\SetMathAlphabet{\mathsfit}{bold}{\encodingdefault}{\sfdefault}{bx}{n}

%% file: sec/0_abstract.tex
\begin{abstract}

Decoding visual representations from brain signals has attracted significant attention in both neuroscience and artificial intelligence. However, the degree to which brain signals truly encode visual information remains unclear. Current visual decoding approaches explore various brain–image alignment strategies, yet most emphasize high-level semantic features while neglecting pixel-level details, thereby limiting our understanding of the human visual system.
In this paper, we propose a brain–image alignment strategy that leverages multiple pre-trained visual encoders with distinct inductive biases to capture hierarchical and multiscale visual representations, while employing a contrastive learning objective to achieve effective alignment between brain signals and visual embeddings. Furthermore, we introduce a Fusion Prior, which learns a stable mapping on large-scale visual data and subsequently matches brain features to this pre-trained prior, thereby enhancing distributional consistency across modalities. Extensive quantitative and qualitative experiments demonstrate that our method achieves a favorable balance between retrieval accuracy and reconstruction fidelity.

\end{abstract}

%% file: sec/1_introduction.tex
\section{Introduction}
With the rapid development of text-to-image generative models \citep{rombach2022high, zhang2023adding, esser2024scaling}, reconstructing human visual stimuli from brain signals has become a prominent research focus in both neuroscience and artificial intelligence. Visual processing is a core function of the human brain. When visual stimuli is processed by the brain, the primary visual cortex initially deciphers basic pixel attributes such as color, edges, and textures, subsequently forwarding them to various higher-order visual cortices for further hierarchical processing \citep{blasdel1983termination, tsumoto1978functional}. These higher-level regions collaborate to synthesize and generalize visual data, resulting in semantic characteristics such as objects and environments, and thus formulating the essential processes underlying human visual perception of the external world. \citep{merigan1993parallel}. 

To investigate these complex and dynamic relationships between the human visual system and brain representations, researchers commonly employ Functional magnetic resonance imaging (fMRI),  Magnetoencephalography (MEG), and Electroencephalogram (EEG) for visual decoding and reconstruction \citep{zhang2025cognitioncapturer, benchetrit2023brain}. fMRI measures brain activity indirectly through blood-oxygen-level-dependent signals, offering high spatial resolution but limited temporal resolution, which makes it difficult to capture rapid neural dynamics \citep{logothetis2001neurophysiological}. In contrast, EEG and MEG directly reflect the brain’s electrophysiological activity. EEG provides high temporal resolution but suffers from low spatial resolution and a poor signal-to-noise ratio. MEG, while also offering millisecond-level temporal precision, provides comparatively better spatial resolution \citep{liu2023emotion, da2013eeg}.

Previous research has explored decoding brain signals by aligning them with visual representations, enabling classification, retrieval, and reconstruction. \citet{song2024decoding} employed contrastive learning to maximize the similarity of matched brain–image pairs while minimizing that of mismatched ones. \citet{li2024visual} proposed the Adaptive Thinking Mapper (ATM) to align brain signal features with CLIP-derived visual embeddings, combined with a two-stage multi-pipe strategy for brain-to-image generation. However, these approaches rely on direct alignment between brain signals and image features, whereas the structural gap between the two modalities makes this strategy insufficient to capture the underlying shared representations.

Recently, several studies have attempted to improve direct alignment by introducing priors or enriching visual representations. \citet{wu2025bridging} introduced the Uncertainty-aware Blur Prior (UBP), which mitigates brain–image mismatches by blurring high-frequency image details. \citet{zhang2025cognitioncapturer} extended CLIP-derived image embeddings with depth information to enhance brain–image alignment. However, these methods focus primarily on high-level semantic alignment while overlooking low-level pixel information. This oversight prevents a comprehensive understanding of the visual content encoded in brain signals and reduces interpretability.

To bridge the structural gap between the temporal dynamics of brain signals and the hierarchical organization of visual representations, we introduce a Hierarchical Visual Fusion framework with a Fusion Prior, inspired by perceptual mechanisms in the human visual system.
The framework integrates multiple pre-trained encoders to construct multiscale visual representations, ranging from pixel-level details to high-level semantics, and leverages contrastive learning to align brain and visual features. To address the limitations of CLIP and related encoders in capturing local and fine-grained information, we incorporate low-level visual features modeled by a Variational Autoencoder (VAE) into the fused representation. 
This fusion substantially improves zero-shot retrieval performance.
In addition, we pretrain a Fusion Prior on large-scale visual data to provide a stable mapping from fused features to diffusion conditions.
Aligning brain embeddings with this prior enables faithful image reconstruction and serves as an effective bridge for a brain-to-image decoding framework.

% \begin{itemize}
% \item We incorporate low-level visual information from a VAE upon semantic alignment, compensating for the limitations of CLIP-based encoders in modeling pixel-level details.

% \item We propose a Fusion Prior that learns a robust visual representation from large-scale data, providing a stable bridge for aligning brain signals to improve cross-modal consistency.

% \item Our method achieves the state-of-the-art performance in retrieval tasks with significant advancements over prior work, while delivering superior reconstruction quality.
% \end{itemize}

\begin{itemize}
    \item We propose a fusion-based brain--vision interface that aligns brain embeddings to a fused visual token built from hierarchical encoders (semantics and pixels), together with a pretrained Fusion Prior that bridges this token to a frozen image generation backbone in a reusable, text-free way.

    \item We provide a key scientific finding: EEG/MEG signals jointly align with high-level semantics and low-level visual details. Within our fusion framework, adding a VAE latent on top of semantic encoders consistently boosts decoding performance, whereas semantics-only or pixels-only settings cannot recover the same brain--vision structure.

    \item Our method achieves state-of-the-art zero-shot retrieval and improved reconstruction quality, while remaining plug-and-play across different brain encoders under a fixed fusion-based training scheme.
    
\end{itemize}

%% file: sec/2_related_work.tex
\section{Related Work}
\label{headings}

% First level headings are in small caps,
% flush left and in point size 12. One line space before the first level
% heading and 1/2~line space after the first level heading.
\paragraph{Brain Visual Decoding} % big backend

% Neural decoding seeks to infer human cognitive and perceptual states from brain signals such as EEG \citep{bai2023dreamdiffusion, li2024visual}, MEG \citep{cichy2016similarity}, or fMRI \citep{kay2008identifying, takagi2023improving}. Within this domain, visual decoding has emerged as one of the most challenging and promising directions, with core tasks including image retrieval \citep{liu2023brainclip} and image reconstruction \citep{ozcelik2023natural, takagi2023high} from neural activity. 
Neural decoding aims to infer human cognitive and perceptual states from brain signals such as EEG \citep{song2024decoding,bai2023dreamdiffusion, li2024visual}, MEG \citep{cichy2016similarity}, or fMRI \citep{kay2008identifying, takagi2023improving}. Among these, visual decoding has become a particularly challenging and promising direction, mainly including tasks like image classification \citep{xu2024beware} , retrieval \citep{liu2023brainclip} and reconstruction \citep{ozcelik2023natural, takagi2023high}. 
A central focus has been on encoding EEG signals into effective representation vectors that capture temporal and frequency characteristics \citep{fu2025brainvis}. To bridge the modality gap, CLIP-based models are commonly adopted as benchmarks \citep{liu2023brainclip, wang2024mindbridge}, while recent methods have explored strategies such as diffusion priors \citep{aggarwal2023controlled} for enhancing semantic consistency in the generative space \citep{ozcelik2023natural, takagi2023improving, li2025neuraldiffuser}, or bidirectional mappings to enforce cross-modal cycle consistency \citep{wei2024mb2c}. At the same time, research on the image modality itself has explored ways to complement the limited semantic expressiveness of neural signals, including blurred preprocessing to suppress high-frequency noise \citep{li2024visual} and textual descriptions to enrich semantic guidance \citep{takagi2023improving}. However, most existing approaches primarily emphasize high-level semantics without sufficiently capturing pixel-level, fine-grained representations, leaving notable gaps in the fidelity of generated or reconstructed images.

\paragraph{Hierarchical and Multiscale Visual Representations}
Recent advances in image-only representation learning emphasize multi-level semantics and dense structure within a single modality. Vision Transformers~\citep{dosovitskiy2020image} trained with self-supervision yield strong global semantics without textual supervision \citep{bao2021beit,xie2022simmim,he2022masked,caron2021emerging,oquab2023dinov2}, while generative latent models such as VAEs and VQ-VAEs provide compact pixel-level codes with high reconstruction fidelity~\citep{kingma2013auto,higgins2017beta,van2017neural,razavi2019generating}. A complementary perspective from neuroscience links deeper network features to higher visual areas and early layers to fine spatial detail and rapid dynamics~\citep{yamins2014performance,cichy2016comparison}, motivating the combination of coarse semantic abstractions with fine-grained local cues. In practice, however, many decoding pipelines \citep{li2024visual,wu2025bridging,zhang2025cognitioncapturer} instantiate a single semantic embedding space for simplicity and zero-shot transfer, which may underweight local structures that are important for faithful image reconstruction from neural signals.

\paragraph{Cross-modal Contrastive Learning}
Cross-modal contrastive learning (CMCL) aligns heterogeneous inputs within a shared embedding space by maximizing agreement between matched pairs under a temperature-scaled InfoNCE objective \citep{oord2018representation,wu2018unsupervised,he2020momentum}. Bi-encoder formulations with cosine similarity and (often) symmetric losses have become the default recipe for scalable pretraining and zero-shot transfer \citep{radford2021learning,jia2021scaling,zhai2022lit,li2022blip}. Building on this recipe, large-scale vision–language systems such as CLIP/ALIGN demonstrate strong generalization across retrieval and classification benchmarks, and the paradigm extends beyond image–text to audio–visual \citep{arandjelovic2017look,morgado2021audio,wu2022wav2clip}, video–language \citep{miech2019howto100m,miech2020end,xu2021videoclip,bain2021frozen,luo2022clip4clip}, and 3D–language \citep{xue2023ulip,zhang2022pointclip,liu2023openshape} alignment. Despite this progress, CMCL typically assumes accurately paired data. At web scale, weak captions, temporal asynchrony, and domain shift impair alignment quality, motivating data curation, caption bootstrapping, and bridging/distillation strategies~\citep{jia2021scaling,miech2019howto100m}. Recent analyses~\citep{liang2022mind,wang2023connecting} also reveal a modality gap between modalities in the shared space, which can complicate fine-grained alignment. In neural decoding, the brain–vision pairing is intrinsically scarce and noisy (limited trials, low SNR, trial-to-trial latency variability), so naively aligning brain signals to a single semantic-only space risks under-representing pixel-level structure and amplifying modality mismatch \citep{cichy2016comparison}.

\paragraph{Adapters for Image Diffusion Models}
Adapters have emerged as parameter-efficient modules that extend pretrained diffusion models with controllability and editing while largely freezing base weights, offering a unifying recipe across tasks and modalities \citep{wang2025image}. T2I-Adapter \citep{mou2024t2i} learns lightweight branches that align external control signals (e.g., edges, depth, sketches) with internal features of a frozen text-to-image model, enabling accurate and composable multi-condition control. ControlNet \citep{zhang2023adding} freezes the pretrained backbone and adds zero-initialized side networks to inject spatial conditions without destabilizing the original prior. IP-Adapter \citep{ye2023ip} decouples cross-attention to integrate image prompts alongside text, delivering strong multimodal conditioning with 22M trainable parameters while keeping the diffusion backbone frozen.

%% file: sec/3_method.tex
\vspace{-0.3cm}
\section{Method}
\label{headings}

% First level headings are in small caps,
% flush left and in point size 12. One line space before the first level
% heading and 1/2~line space after the first level heading.
\vspace{-0.2cm}
\subsection{Problem Statement}

% \begin{figure}[htbp]
%     \centering
%     \includegraphics[width=\linewidth]{figs/structure.png}
%     \caption{Overview of the retrieval and generation pipeline and our model framework.}
%     \label{fig:structure}
% \end{figure}
\begin{figure}[htbp]
  \centering
  \begin{subfigure}[t]{0.48\textwidth}
    \centering
    \includegraphics[width=\linewidth]{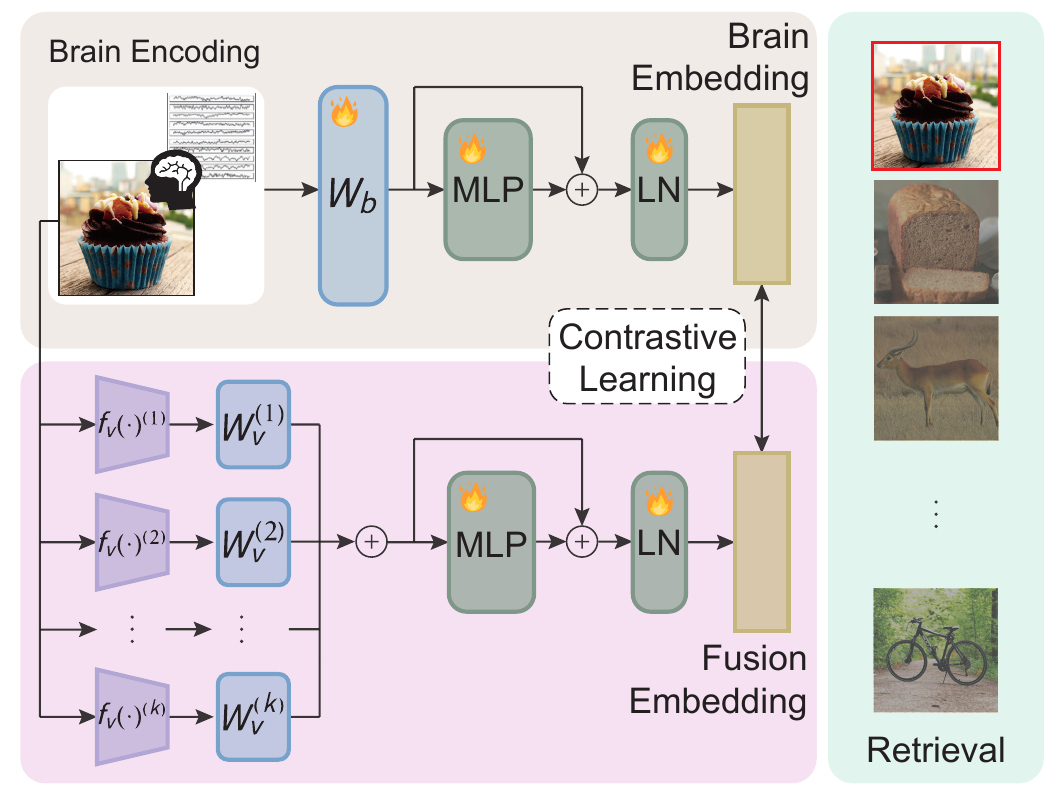}
    \caption{Brain-to-image retrieval.}
    \label{fig:sub1}
  \end{subfigure}%
  % split line
  \hfill
  {\color{gray!40}\vrule width 0.5pt} % 
  \hfill
  \begin{subfigure}[t]{0.48\textwidth}
    \centering
    \includegraphics[width=\linewidth]{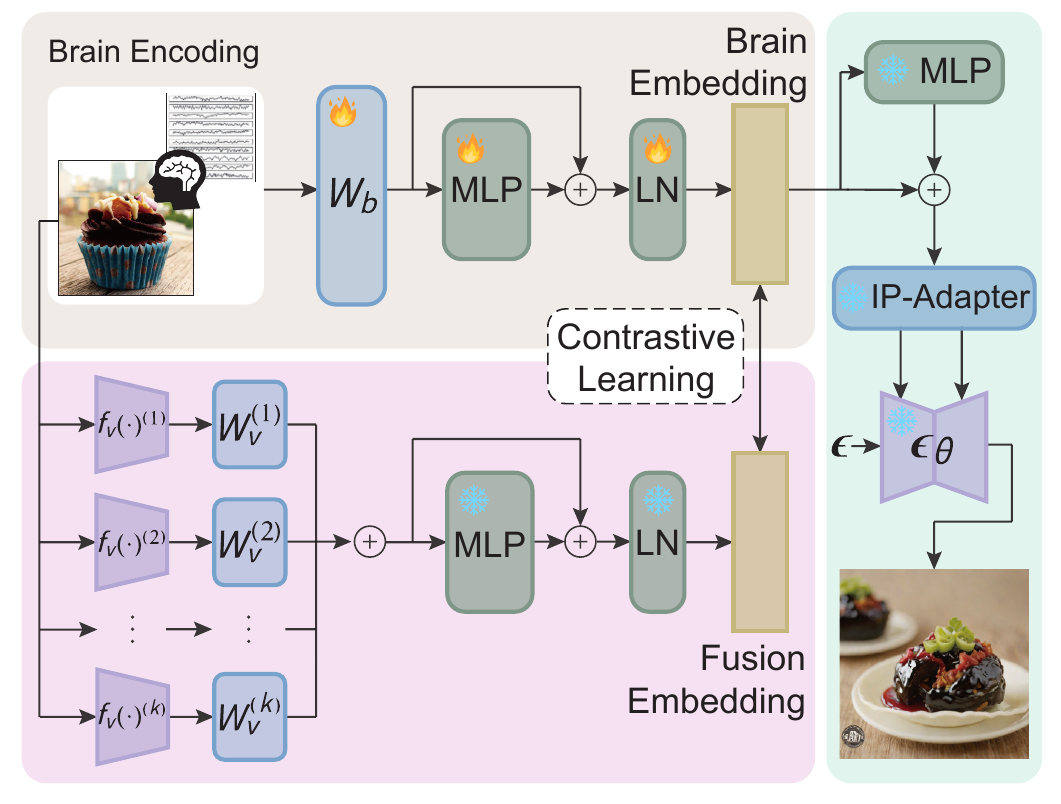}
    \caption{Brain-to-image reconstruction.}
    \label{fig:sub2}
  \end{subfigure}
  \caption{\textbf{Learning pipelines}. Left: Retrieval objective that aligns the brain embedding $z_b$ with the fused visual embedding $z_f$ (HVF over $K$ pretrained encoders) using a symmetric InfoNCE; evaluation is nearest-neighbor retrieval in the fused space. Right: Reconstruction pipeline with a frozen, pretrained fusion prior—HVF plus a Conditioning Adapter (MLP projector + IP-Adapter with decoupled cross-attention). We contrastively align $z_b$ to the frozen $z_f$, project to $z_c$, and inject $z_c$ into a frozen SDXL UNet to synthesize the image. Visual encoders and the UNet are frozen; only the brain side is updated during alignment.}
  \label{fig:structure}
\end{figure}

% In the brain visual decoding task, our goal is to retrieve or reconstruct the corresponding visual content from recorded brain signals. Let the paired brain signals and visual images be $(x_b, x_v) \in \mathcal{D}$, where $x_v \in \mathbb{R}^{H \times W \times 3}$ represents the visual stimulus image, and $x_b \in \mathbb{R}^{C \times T}$ denotes the high-temporal-resolution brain signal captured under the same stimulus. The corresponding marginal distributions are denoted as $\mathcal{P}_v$ and $\mathcal{P}_b$ for visual and brain modalities, respectively.

The goal of visual decoding is to retrieve or reconstruct the visual information corresponding to recorded brain signals. We denote paired brain signals and visual images as $(x_v, x_b) \in \mathcal{D}$, where $x_v \in \mathbb{R}^{H \times W \times 3}$ represents the visual stimulus, with $H$ and $W$ denoting the image height and width, respectively. $x_b \in \mathbb{R}^{C \times T}$ represents the brain signals recorded under the same stimulus, where $C$ corresponds to the number of electrode channels and $T$ indicates the length of time samples. 
% The marginal distributions are denoted as $\mathcal{P}_v$ and $\mathcal{P}_b$ for the visual and brain modalities, respectively.

\subsection{Aligning Brain Signals with Hierarchical Visual Representations}
\label{align-sec}

% overview
% Directly aligning brain signals with a single visual encoding often fails to capture multi-scale information inherent in the visual modality. High-level semantic features are critical for category recognition, while low-level features provide complementary structural and pixel-level details. Our solution is to construct a unified hierarchical visual representation by integrating multiple pre-trained visual encoders and aligning it with brain signal embeddings via a contrastive learning objective.

% lack of reference TODO
Directly aligning brain signals with visual representations may fail to capture the intrinsic multiscale nature of the visual information, thereby limiting alignment performance. While high-level semantic features in the visual modality are crucial for category recognition and abstract understanding, low-level features provide complementary structural information and pixel-level details, which are indispensable for improving reconstruction quality. Inspired by this visual perception mechanism \citep{blasdel1983termination, tsumoto1978functional}, we integrate multiple pretrained visual encoders to separately extract high-level semantic features and low-level pixel features, and align them with brain signal embeddings through a contrastive learning objective to construct a unified hierarchical visual representation.

\paragraph{Hierarchical visual representations}
% To obtain rich and complementary multiscale visual features, we extract embeddings with $K$ (we set $K=3$ by default) pre-trained encoders $f_v^{(k)}$ applied to the image $x_v$, yielding $z_v^{(k)} = f_v^{(k)}(x_v)$ for $k{=}1,\dots,K$.  
% Specifically, we integrate multiple CLIP encoders to capture high-level semantic features spanning both global and fine-grained levels. Building on this, we incorporate a VAE to encode low-level pixel details, offering complementary cues beyond high-level semantics. The VAE produces a latent tensor of shape $[H/8, W/8, 4]$, which is flattened into a vector of dimension $(H/8)(W/8)\!\times\!4 = HW/16$, thereby retaining local structures and visual details. 

As depicted in Fig.~\ref{fig:structure}-(a), we devise a multi-head encoder structure to obtain hierarchical visual representations ranging from high-level visual semantics (e.g., objects, scenes, and relations) to low-level visual features (e.g., colors, textures, and layouts). We apply $K$ pretrained encoders ($K{=}3$ by default) to the image $x_v$, yielding $z_v^{(k)} = f_v^{(k)}(x_v)$ for $k{=}1,\dots,K$. For high-level visual semantics, we integrate multiple CLIP encoders and use a single global token from each (i.e.,  [CLS] token for ViT-based models and the pooled projection for ResNet-based models). For low-level visual features, the VAE encoder outputs a latent of shape $[H/8, W/8, 4]$, which we flatten into a vector of length $(H/8)(W/8)\!\times\!4 = HW/16$, preserving local structure and visual detail.

% Then, each $z_v^{(k)}\in\mathbb{R}^{d_k}$ is projected to a shared dimension $d$ via a learned linear map $W^{(k)}\in\mathbb{R}^{d_k\times d}$, formulated as:

% \begin{equation}
% \tilde{z}_v^{(k)} = z_v^{(k)} W^{(k)},
% \end{equation}

% Subsequently, we introduce a post-norm residual Hierarchical Visual Fuser (HVF) to integrate features from multiple encoders, yielding a representation that captures both global semantics and pixel-level details. This process is defined as:
% \begin{equation} 
%     \label{fusion} 
%     z_f = \mathrm{LayerNorm}\Big(\sum_{k=1}^{K} \tilde{z}_v^{(k)} + \phi(\sum_{k=1}^{K} \tilde{z}_v^{(k)})\Big)=\mathrm{HVF}(\{\tilde{z}_v^{(k)}\}_{k=1}^{K}), 
% \end{equation} 
% where $\phi$ is a two-layer MLP with hidden size $d_m$ and GELU activation.

We fuse features with a post-norm residual Hierarchical Visual Fuser (HVF). For each encoder, a learned linear map $W_v^{(k)}\!\in\!\mathbb{R}^{d_k\times d}$ aligns the embedding to the shared dimension $d{=}1024$:
\begin{equation}
\begin{aligned}
\bar{z}_v = \sum_{k=1}^{K} z_v^{(k)} W_v^{(k)},
\end{aligned}
\end{equation}
The aligned features are fused with a residual Multi-Layer Perceptron (MLP), and we have
\begin{equation}
\begin{aligned}
z_f = \mathrm{LayerNorm}\big(\bar{z}_v + \phi_v(\bar{z}_v)\big)
% =\mathrm{HVF}\left({z_v^{(k)}}_{k=1}^{K}\right)
,
\label{eq:fusion}
\end{aligned}
\end{equation}

where $\phi_v$ denotes a two-layer MLP with hidden size $d_v = 1024$ and GELU activation.

\paragraph{Contrastive learning objective}
% For the brain signal modality, we use a lightweight brain projection network $f_b(\cdot)$ as proposed in \citep{wu2025bridging} to extract latent brain embeddings in the same $d$-dimensional space as $z_f$, so we have

% For the brain signal modality, we adopt a lightweight brain projection network $f_b(\cdot)$ to map brain signals into the same $d$-dimensional space as $z_f$, yielding

% \begin{equation}
%     z_b = f_b(x_b) \in \mathbb{R}^d.
% \end{equation}

For the brain modality, we adopt an MLP-based Brain Projection (MBP) network that projects the EEG signal to an embedding. We first flatten the preprocessed signal to $x_b'\in\mathbb{R}^{(C\cdot T)}$, then align it to the visual embedding width using a learned linear projection $W_b \in \mathbb{R}^{CT \times d}$. We then reuse the same architecture as Eq.~(\ref{eq:fusion}) to produce a $d$-dimensional embedding compatible with $z_f$ with a hidden size of $d$ that
\begin{equation}
    \begin{aligned}
        \bar{z}_b = x_b' W_b,\qquad
 z_b = \mathrm{LayerNorm}\big(\bar{ z}_b + \phi_b(\bar{ z}_b)\big),
    \end{aligned}
\end{equation}
where $\phi_b$ denotes a two-layer MLP with hidden size $d_b = 1024$ and GELU activation.

% We adopt a CLIP-like \citep{radford2021learning} contrastive loss (i.e., the InfoNCE loss \citep{oord2018representation}) to align $z_b$ with the fused visual representation $z_f$. For a batch of size $N$, the objective is

% \begin{equation}
%     \mathcal{L}_\text{contrastive} = -\frac{1}{2N}[ \sum_{i=1}^N\log \frac{\exp(z_b^{(i)}\cdot z_\text{m}^{(i)} / \tau)}{\sum_{j=1}^N \exp(z_b^{(j)}\cdot z_\text{m}^{(j)} / \tau )}+ \sum_{j=1}^N\log \frac{\exp(z_b^{(j)}\cdot z_\text{m}^{(j)} / \tau)}{\sum_{j=1}^N \exp(z_b^{(i)}\cdot z_\text{m}^{(i)} / \tau)}].
% \end{equation}

% where $\tau$ is a trainable temperature parameter. Learning this objective encourages each brain signal embedding to be closest to its corresponding visual embedding within the batch.

% Following \citet{radford2021learning}, we adopt a CLIP-style InfoNCE loss~\citep{oord2018representation}. With $N$ paired samples, we compute cosine-similarity logits with a trainable temperature $\tau$

We employ a CLIP-style InfoNCE loss~\citep{oord2018representation} to align brain and visual embeddings. Given $N$ paired samples, we compute cosine-similarity logits with a trainable temperature $\tau$ (initialized to 0.07):
\begin{equation}
    \hat z_b^{(i)} = \frac{z_b^{(i)}}{\|z_b^{(i)}\|_2},\quad
    \hat z_f^{(i)} = \frac{z_f^{(i)}}{\|z_f^{(i)}\|_2},\quad
    s_{ij} = \frac{\hat z_b^{(i)\top}\hat z_f^{(j)}}{\tau},
\end{equation}
where $\|\cdot\|_2$ is L2 norm. The  learning objective is defined as
\begin{equation}
\label{constrastive}
\mathcal{L}_\text{contrastive}
= -\frac{1}{2N}\Bigg(
\sum_{i=1}^{N}\log\frac{\exp(s_{ii})}{\sum_{j=1}^{N}\exp(s_{ij})}
+
\sum_{i=1}^{N}\log\frac{\exp(s_{ii})}{\sum_{j=1}^{N}\exp(s_{ji})}
\Bigg),
\end{equation}
% which encourages each brain embedding to align most strongly with its paired fused-visual embedding within the batch.
\subsection{Pretrained Fusion Prior for Reconstruction}
% section overview
While the above contrastive learning aligns brain signals with hierarchical visual representations, directly feeding these fused representations into a pretrained diffusion model for reconstruction often results in unstable outputs. 
The core issue is the absence of a stable conditioning prior: brain-driven features do not yet match the distribution expected by the generative model, leading to noisy or misaligned guidance. To address this, we introduce the fusion prior to learn a robust mapping from fused visual features to diffusion conditions.
% Second level headings are in small caps,
% flush left and in point size 10. One line space before the second level
% heading and 1/2~line space after the second level heading.

\paragraph{Fusion prior pretraining}
% As depicted in Fig.~\ref{fig:structure}, we pass the fused feature $z_f$ into a Conditioning Adapter (CA), an MLP projector followed by an IP-Adapter with decoupled cross-attention. The projector produces 
As depicted in Fig.~\ref{fig:structure}-(b), we first feed the fused visual representation $z_f$ from the HVF into an additional projector to obtain $z_c$:
\begin{equation}
\label{eq:proj}
    z_c = z_f + \phi_c(z_f), 
\end{equation}
where $z_f,z_c\in\mathbb{R}^d$ and both $\phi_v$ in Eq.~\ref{eq:fusion} and $\phi_c$ denotes a two-layer MLP with hidden size $d_c = 4096$ and GELU activation. The IP-Adapter~\citep{ye2023ip} then injects $z_c$ into a frozen SDXL~\citep{podell2023sdxl} UNet via cross-attention. 
% We denote the overall conditioning path compactly as $\mathrm{CA}(z_f)$.
% These features are then injected into a pretrained text-to-image diffusion model (SDXL \citep{podell2023sdxl}) via an IP-Adapter~\citep{ye2023ip} with decoupled cross-attention. We denote the overall conditioning path compactly as $\text{CA}(z_f)$.
Given noisy latent $x_t$ at timestep $t$, the whole network ${\delta}$ is trained to predict the noise $\epsilon$ with
\begin{equation}
\label{prior}
    \mathcal{L}_\text{prior} = \parallel \epsilon - {\delta}(x_t, t, z_c) \parallel_2^2,
\end{equation}
where $\epsilon\sim\mathcal{N}(0,I)$ is the diffusion target and $\mathcal{L}_\text{prior}$ is the loss function.  

During pretraining on large-scale visual data, the UNet backbone remains frozen, while the HVF and the projector are trained from scratch, the IP-Adapter is initialized from pretrained weights\footnote{\url{https://huggingface.co/h94/IP-Adapter/resolve/main/sdxl_models/ip-adapter_sdxl_vit-h.safetensors}} to accelerate convergence. Text prompts are left empty, ensuring the model learns a text-free mapping from fused visual features to diffusion conditions.

\paragraph{Brain-to-fusion alignment}
% Once the HVF is pretrained, we freeze it and update only the brain encoder using a combined objective that adds an MSE term to the contrastive loss in Eq.~(\ref{constrastive}):
% \[
% \mathcal{L}=\lambda\,\mathcal{L}_\text{contrastive}+(1 - \lambda)\,\mathcal{L}_\text{mse},\quad
% \mathcal{L}_\text{mse}=\frac{1}{N}\sum_{i=1}^{N}\big\|\hat z_b^{(i)}-\hat z_f^{(i)}\big\|_2^2,
% \]
% where $\lambda$ balances the two terms. This joint objective projects brain-derived embeddings into the stable, pretrained fusion space, thereby mitigating representational drift and yielding robust reconstruction when passed to the diffusion model.

Once the HVF is pretrained, we freeze it and update only the brain encoder using the same loss function $\mathcal{L}_\text{contrastive}$ as in Eq.~(\ref{constrastive}),
which ensures that brain-derived embeddings are projected into a stable, pretrained fusion space. This prevents representational drift and yields robust reconstruction when passed to the diffusion model.

\paragraph{Full pipeline for reconstruction}
In all, training uses two stages and inference one. (i) \emph{Prior pretraining:} for input images $x_v$, extract $\{z_v^{(k)}\}_{k=1}^K$, fuse and project them via the HVF and projector to obtain $z_c$, and train the IP-Adapter jointly with the HVF and projector (UNet frozen) by minimizing $\mathcal{L}_\text{prior}$ in Eq.~(\ref{prior}) under empty text prompts, yielding a stable, text-free fusion prior. (ii) \emph{Brain–fusion alignment:} freeze the pretrained fusion prior (HVF, projector and IP-Adapter) and the UNet, and update only the brain side (i.e., the MBP module only) on paired $(x_b,x_v)$ with the symmetric InfoNCE loss in Eq.~(\ref{constrastive}) so that $z_b$ lies in the fusion space of $z_f$. (iii) \emph{Reconstruction:} given test brain signals $x_b$, compute $z_b=f_b(x_b)$, feed it to the projector to obtain $z_c$, and use $z_c$ as the sole condition for the frozen IP-Adapter/UNet; a standard diffusion sampler(SDXL uses an Euler–ancestral sampler~\citep{karras2022elucidating}) then produces $\hat{x}_v$, yielding stable and semantically faithful reconstructions.

%% file: sec/4_experiment.tex
\section{Experiment}
\subsection{Experimental Details}

We train the contrastive stage on a single NVIDIA 5090 32GB GPU for 25 epochs with a global batch size of 1024. We use AdamW with a peak learning rate of $5\times10^{-4}$ under a cosine decay schedule and a 10-step warmup from zero. Unless otherwise stated, retrieval uses a fixed encoder set comprising OpenAI CLIP RN50, LAION CLIP ViT-B/32~\citep{schuhmann2022laion}, and an SDXL VAE; each backbone follows its canonical preprocessing. The VAE supports multiple input resolutions and defaults to $128\times128$. For generation, we swap RN50 for LAION CLIP ViT-H/14, freeze the pretrained HVF on the visual side, and train only the MBM module of the brain modality.
\begin{figure}[htbp]
    \centering
    \includegraphics[width=0.95\linewidth]{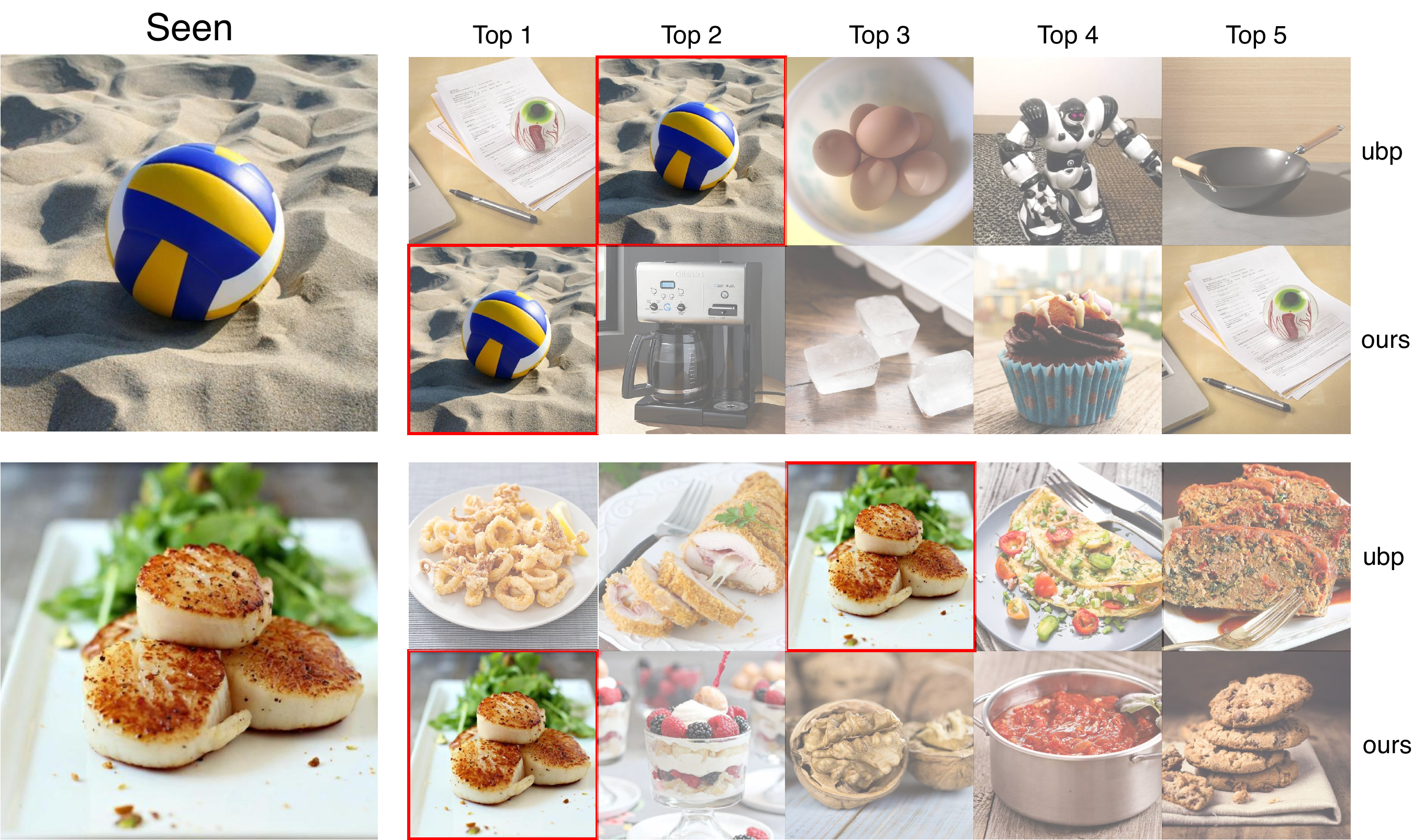}
    \caption{\textbf{Hard-case retrieval comparison.} The top-5 retrieved images on the hard-case set from our method and the UBP baseline.
}
    \label{fig:visual-retrieval}
\end{figure}
\begin{figure}[htbp]
    \centering
    \includegraphics[width=0.95\linewidth]{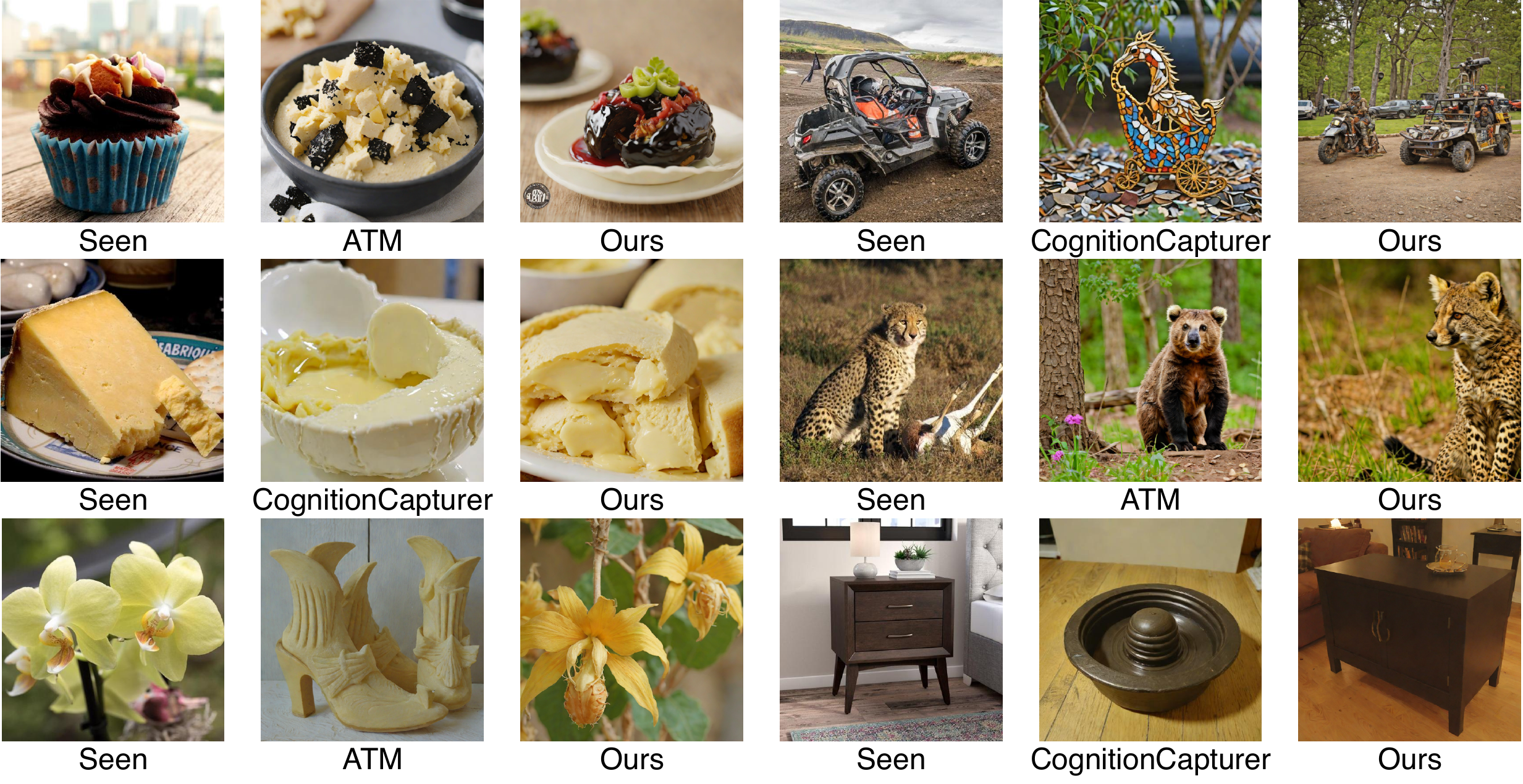}
    \caption{\textbf{Qualitative comparison of brain-to-image reconstructions.} Each triplet shows the ground-truth stimulus (left), baseline (middle), and our reconstruction (right). All examples use EEG recordings from subject 8.
}
    \label{fig:visual-comparison}
\end{figure}
The contrastive stage is trained on THINGS-EEG~\citep{grootswagers2019representational,gifford2022large} and THINGS-MEG~\citep{hebart2023things}. For THINGS-EEG (10 participants), the training split contains 1654 concepts with 10 images per concept and 4 repetitions per image; the test split contains 200 concepts with 1 image per concept and 80 repetitions per image. We follow prior work\citep{li2024visual,wu2025bridging} to select 17 occipito–parietal channels (O+P) and standard preprocessing \citep{song2024decoding}. For THINGS-MEG (4 participants, 271 channels), the training split consists of $1854{\times}12{\times}1$ (concepts × images × reps) and the test split $200{\times}1{\times}12$. To improve signal-to-noise ratio(SNR), repetitions for the same stimulus are averaged within subject in both datasets (training and test). Additional details are provided in the appendix~\ref{appendix:dataset}.
\input{tabs/retri_all_avg}
\input{tabs/recon}
For fusion-prior pretraining, we explore multiple prior configurations. Unless noted, training uses two NVIDIA 5090 32 GB GPUs, a fixed learning rate of $1\times10^{-4}$, SDXL-base as the diffusion backbone, and the largest feasible batch size of 12 per GPU. Each configuration is trained at $512\times512$ for 100k steps, about two epochs, and takes roughly 15 hours per prior configuration. Pretraining uses ImageNet-1k with about 1.3M images. For reconstruction at inference we use SDXL-Turbo with a 4-step sampler for fast evaluation.

\subsection{Quantitative Evaluation}
We evaluate two tasks, brain–visual retrieval and brain–visual reconstruction. For retrieval, we report 200-way zero-shot top-1 and top-5 accuracy on THINGS-EEG and THINGS-MEG under both intra-subject and inter-subject protocols. For reconstruction, following prior work \citep{ozcelik2023natural,benchetrit2023brain,li2024visual}, we measure low-level fidelity with PixCorr and SSIM and adopt the remaining semantic and feature-level metrics from these works, including AlexNet(2/5), Inception, CLIP and SwAV distance, where lower is better.
The retrieval baselines are BraVL~\citep{du2023decoding}, NICE and its spatial variants (NICE-SA, NICE-GA)~\citep{song2024decoding}, ATM~\citep{li2024visual}, VE-SDN~\citep{chen2024visual}, MB2C~\citep{wei2024mb2c}, UBP~\citep{wu2025bridging}, and CognitionCapturer (C.C., All/Image/Depth/Text)~\citep{zhang2025cognitioncapturer}. For reconstruction, we compare with ATM~\citep{li2024visual}, CognitionCapturer~\citep{zhang2025cognitioncapturer}, and Brain Decoding (B.D.)~\citep{benchetrit2023brain}. When prior work reports single-subject results only (e.g., ATM on subj-8), we indicate this in the tables.

% Table~\ref{tab:retrieval} shows that our method sets a new state of the art on both datasets and both protocols. On EEG intra-subject it moves top-1 from the best prior 50.9 with UBP to 73.0 while top-5 rises to 94.1. On EEG inter-subject it improves top-1 from 12.4 with UBP to 23.1 and top-5 from 33.4 to 50.6. On MEG intra-subject it advances top-1 from 26.7 with UBP to 37.8 and top-5 from 55.2 to 67.0. On MEG inter-subject it improves top-1 from 2.2 with UBP to 6.0 and top-5 from 10.4 to 16.8. Our approach achieves most significant gains in the cross-participant regime, indicating stronger generalization rather than only better within-subject fitting. Results are consistent for both top-1 and top-5.

Compared to the strongest prior work (UBP), our model consistently improves 200-way zero-shot retrieval across all protocols (Top-1/Top-5): EEG intra 75.7/94.6 vs 50.9/79.7, EEG inter 20.0/44.1 vs 12.4/33.4, MEG intra 33.7/60.5 vs 26.7/55.2, and MEG inter 5.4/15.2 vs 2.2/10.4 (Tab.~\ref{tab:retrieval}). Gains are largest in the inter-subject setting, indicating stronger cross-participant generalization.

Table~\ref{tab:reconstruction} summarizes reconstruction. On MEG our model matches or exceeds prior work on both low-level similarity and semantic alignment while maintaining a competitive SwAV distance. On EEG it improves the commonly reported subj-8 case and delivers clear subject-averaged gains over ATM and C.C. The average EEG PixCorr increases from 0.150 with C.C.(All) to 0.186 with our model while SSIM remains comparable, and semantic similarities improve across AlexNet, Inception, and CLIP with a lower SwAV distance than C.C. and B.D. Taken together, the metrics indicate that our approach raises both fidelity and semantic agreement and that the improvements persist beyond single-subject evaluation.
\input{tabs/retri_ablation}
% Per-subject analyses for both tasks are provided in the appendix and follow the same trend.

\subsection{Visual Comparison}
In Fig.~\ref{fig:visual-retrieval}, we show the top-5 retrieved images on the Hard-Case set for our method and the UBP baseline, with our method performing better.
We further provide qualitative comparisons with previous brain decoding approaches. As shown in Fig.~\ref{fig:visual-comparison}, our method reconstructs images with clearer object contours and more faithful color distribution compared to CognitionCapturer \citep{zhang2025cognitioncapturer} and ATM \citep{li2024visual}. In particular, our reconstructions preserve fine-grained structural details while capturing semantically consistent attributes that are often missing in the baselines. Moreover, the overall perceptual quality aligns more closely with the ground-truth stimuli, demonstrating the effectiveness of our framework in bridging brain signals and visual representations.
\input{tabs/recon_ablation}
\input{tabs/backbone_eeg}

\subsection{Ablation Studies}
\begin{figure}[htbp]
    \centering
    \includegraphics[width=0.95\linewidth]{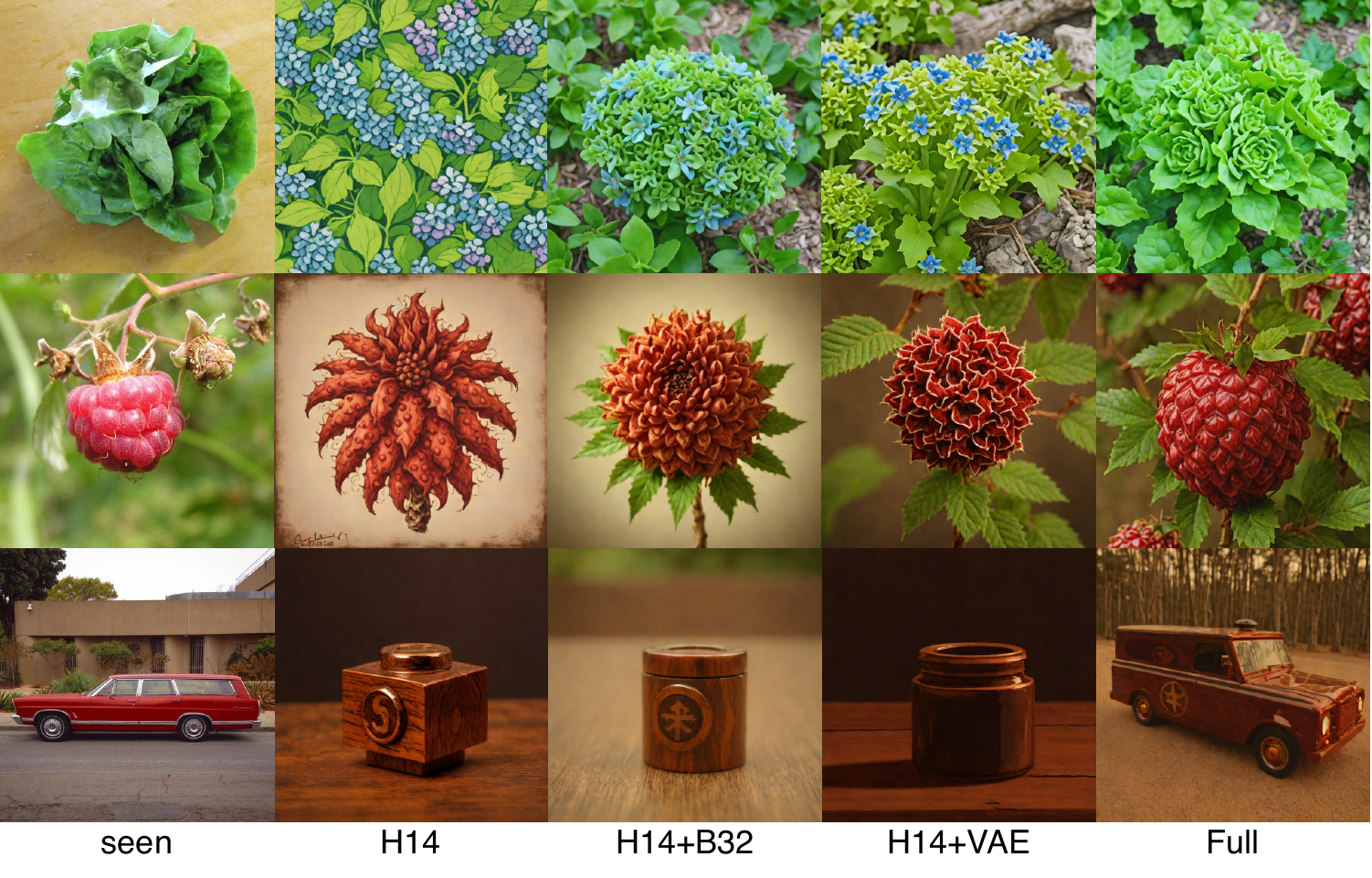}
    \caption{
    % \textcolor{BrickRed}{\textbf{Ablation on fusion priors.} Each row shows the ground-truth stimulus and reconstructions produced with different fused configurations: H14, H14{+}B32, H14{+}VAE, and H14{+}B32{+}VAE. All examples use EEG recordings from subject 8.}
    \textbf{Ablation on fusion priors.} Each row shows the ground-truth stimulus and reconstructions produced with different fused configurations: H14, H14{+}B32, H14{+}VAE, and H14{+}B32{+}VAE. All examples use EEG recordings from subject 8.
    }
    \label{fig:visual-ablation}
\end{figure}
We ablate the visual encoder composition on 200-way zero-shot EEG retrieval in Tab.~\ref{tab:eeg_ablation}. Single encoders provide reasonable baselines (e.g., B32: 52.2/83.3 top-1/top-5 intra-subject; 13.3/33.9 inter-subject), and stacking semantic encoders (RN50+B32) yields modest gains (56.9/86.1 intra; 14.4/36.8 inter). Adding the VAE latent gives the largest improvements: B32+VAE reaches 73.6/94.3 (intra) and 19.1/41.2 (inter), and RN50+VAE reaches 65.8/90.4 (intra) and 17.4/37.3 (inter). The full RN50+B32+VAE fusion achieves 75.7/94.6 (intra) and 20.0/44.1 (inter), giving a small but consistent boost over the best pairwise settings on both intra- and inter-subject splits.
\begin{figure}[t]
\centering

\begin{subfigure}[t]{0.45\linewidth}
  \centering
  \includegraphics[width=\linewidth]{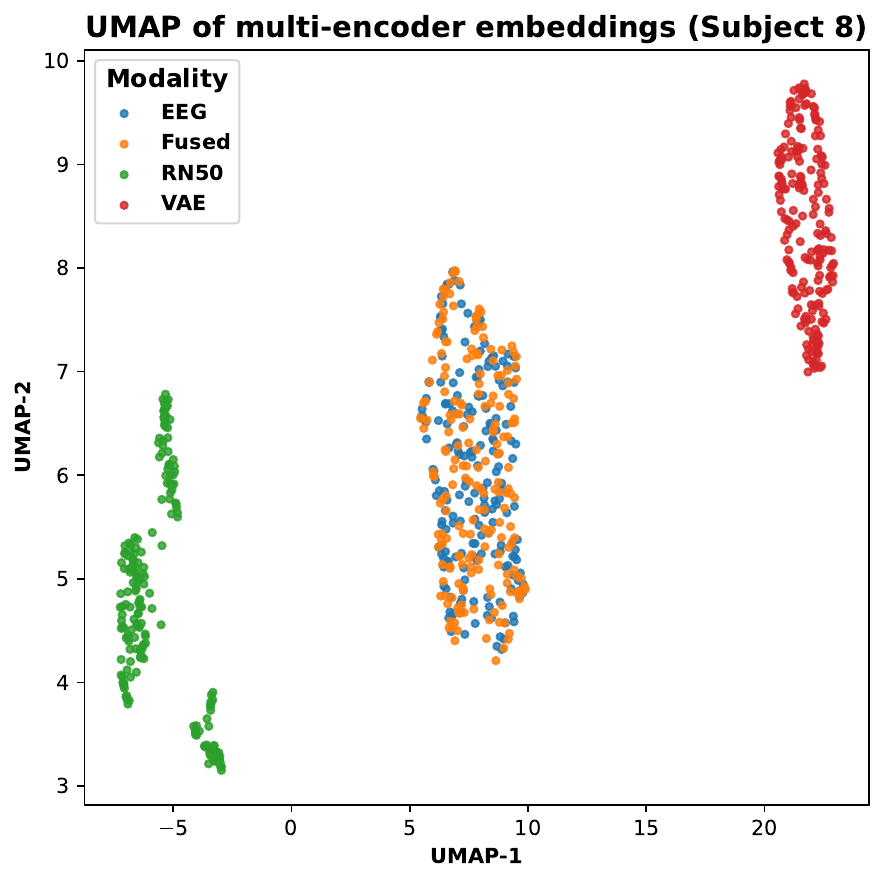}%
  \label{fig:umap_multi_encoder_subj8}
\end{subfigure}
\hfill
\begin{subfigure}[t]{0.45\linewidth}
  \centering
  \includegraphics[width=\linewidth]{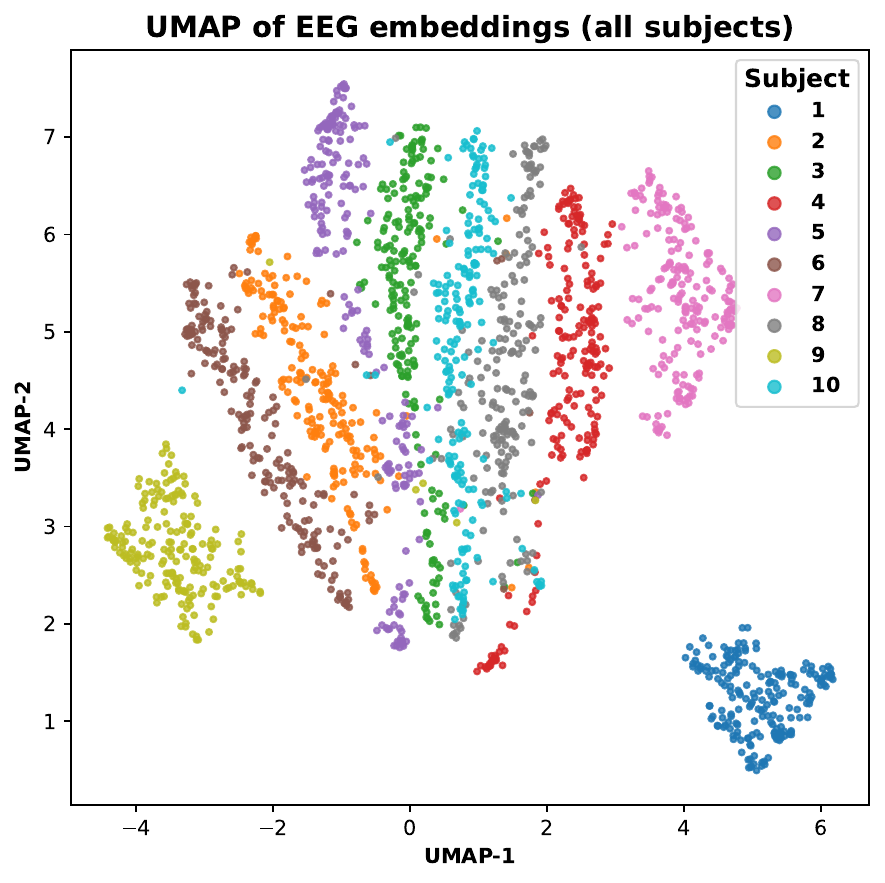}%
  \label{fig:umap_eeg_all_subjects}
\end{subfigure}

\caption{
% \textcolor{Purple}{UMAP visualization of learned embeddings on the test split of the THINGS-EEG dataset.
% Left: multi-encoder visual embeddings (RN50, flattened VAE) and the fused token are projected together with Subject~8 EEG embeddings.
% Right: EEG embeddings from all 10 subjects.}
UMAP visualization of learned embeddings on the test split of the THINGS-EEG dataset.
Left: multi-encoder visual embeddings (RN50, flattened VAE) and the fused token are projected together with Subject~8 EEG embeddings.
Right: EEG embeddings from all 10 subjects.
}
\label{fig:umap_overview}
\end{figure}
For brain-to-image reconstruction, Tab.~\ref{tab:ablation_reconstruction} compares fusion priors built from H14, B32, and VAE features. Adding B32 to H14 improves both low-level (PixCorr, SSIM) and high-level perceptual metrics, and the full H14+B32+VAE prior further boosts PixCorr and AlexNet similarities while keeping the remaining metrics close to the best two-stream setting; \textcolor{BrickRed}{Fig.~\ref{fig:visual-ablation}} shows representative qualitative differences between these configurations.

% \textcolor{BurntOrange}{Tab.~\ref{tab:backbone_eeg} reports the backbone comparison under the same 17-channel O+P setting and the same fusion-based training. We plug ShallowNet~\citep{schirrmeister2017deep}, DeepNet~\citep{schirrmeister2017deep}, EEGNet~\citep{lawhern2018eegnet}, TSConv~\citep{song2024decoding}, ATM~\citep{li2024visual}, BrainProjection~\citep{wu2025bridging} and our MBP into the fused visual interface (RN50, B32, VAE and their combinations), keeping the visual encoders, alignment objective, and loss fixed. All prior backbones obtain higher top-1/top-5 retrieval accuracy than their original single-encoder reports, and MBP achieves the highest accuracy in every visual-encoder combination; in the Full RN50+B32+VAE configuration, it reaches 75.7\%/94.6\% compared to 72.9\%/94.1\% for BrainProjection, showing plug-and-play use of the same fusion interface across EEG backbones.}
Tab.~\ref{tab:backbone_eeg} reports the backbone comparison under the same 17-channel O+P setting and the same fusion-based training. We plug ShallowNet~\citep{schirrmeister2017deep}, DeepNet~\citep{schirrmeister2017deep}, EEGNet~\citep{lawhern2018eegnet}, TSConv~\citep{song2024decoding}, ATM~\citep{li2024visual}, BrainProjection~\citep{wu2025bridging} and our MBP into the fused visual interface (RN50, B32, VAE and their combinations), keeping the visual encoders, alignment objective, and loss fixed. All prior backbones obtain higher top-1/top-5 retrieval accuracy than their original single-encoder reports, and MBP achieves the highest accuracy in every visual-encoder combination; in the Full RN50+B32+VAE configuration, it reaches 75.7\%/94.6\% compared to 72.9\%/94.1\% for BrainProjection, showing plug-and-play use of the same fusion interface across EEG backbones.

% \textcolor{Purple}{\subsection{Qualitative analysis of learned embeddings}
% Fig.~\ref{fig:umap_overview} visualizes the learned embeddings, evaluated on the THINGS-EEG test split after training on the training split. In the left panel, Subject~8 EEG embeddings lie very close to the fused visual tokens and far from both RN50 and VAE embeddings, showing that EEG is aligned to the fused representation rather than to any single encoder. In the right panel, EEG embeddings from the 10 subjects form clearly separated clusters, indicating subject-specific semantic distributions that match the strong inter-subject variability typically observed in EEG. Fig.~\ref{fig:similarity_eeg_image_subj8} (in Appendix) exhibits a sharp diagonal between EEG and image embeddings, confirming that the learned fused representation supports reliable 200-way zero-shot retrieval on the test set.}
\subsection{Qualitative analysis of learned embeddings}
Fig.~\ref{fig:umap_overview} visualizes the learned embeddings, evaluated on the THINGS-EEG test split after training on the training split. In the left panel, Subject~8 EEG embeddings lie very close to the fused visual tokens and far from both RN50 and VAE embeddings, showing that EEG is aligned to the fused representation rather than to any single encoder. In the right panel, EEG embeddings from the 10 subjects form clearly separated clusters, indicating subject-specific semantic distributions that match the strong inter-subject variability typically observed in EEG. Fig.~\ref{fig:similarity_eeg_image_subj8} (in Appendix) exhibits a sharp diagonal between EEG and image embeddings, confirming that the learned fused representation supports reliable 200-way zero-shot retrieval on the test set.

%% file: tabs/retri_all_avg.tex
\begin{table*}[t]
\centering
\caption{Average Top-1 / Top-5 accuracy (\%) for 200-way zero-shot retrieval on THINGS-\textbf{EEG} and THINGS-\textbf{MEG}. 
All numbers are subject-wise averages; “–” indicates not reported.}
\label{tab:retrieval}
\footnotesize
\setlength{\tabcolsep}{7pt}
\renewcommand{\arraystretch}{1.08}
\begin{tabular}{l cc cc cc cc}
\toprule
\multirow{3}{*}{\textbf{Method}}
& \multicolumn{4}{c}{\textbf{EEG}} & \multicolumn{4}{c}{\textbf{MEG}} \\
\cmidrule(lr){2-5}\cmidrule(lr){6-9}
& \multicolumn{2}{c}{\textbf{Intra-subject}} & \multicolumn{2}{c}{\textbf{Inter-subject}}
& \multicolumn{2}{c}{\textbf{Intra-subject}} & \multicolumn{2}{c}{\textbf{Inter-subject}} \\
& Top-1 & Top-5 & Top-1 & Top-5 & Top-1 & Top-5 & Top-1 & Top-5 \\
\midrule
BraVL    & 5.8  & 17.5 & 1.8  & 7.0  & –    & –    & –    & –    \\
NICE     & 16.1 & 43.6 & 6.2  & 21.4 & 12.8 & 36.0 & –    & –    \\
NICE-SA  & 14.7 & 41.7 & 7.0  & 23.1 & 12.7 & 35.0 & –    & –    \\
NICE-GA  & 15.6 & 42.8 & 5.9  & 21.6 & 14.3 & 42.3 & –    & –    \\
MB2C     & 28.5 & 60.4 & 11.9   & 32.0    & –    & –    & –    & –    \\
ATM      & 28.5 & 60.4 & 11.8 & 33.7 & -    & -    & –    & –    \\
VE-SDN   & 37.2 & 69.9 & –    & –    & –    & –    & –    & –    \\
CC-All   & 35.6 & 80.2 & –    & –    & –    & –    & –    & –    \\
UBP      & 50.9 & 79.7 & 12.4 & 33.4 & 26.7 & 55.2 & 2.2  & 10.4 \\
\textbf{Ours} & \textbf{75.7} & \textbf{94.6} & \textbf{20.0} & \textbf{44.1} & \textbf{33.7} & \textbf{60.5} & \textbf{5.4} & \textbf{15.2} \\
\bottomrule
\end{tabular}
\end{table*}

%% file: tabs/recon.tex
\begin{table}[h]
\centering
\caption{Quantitative assessments of the reconstruction quality for EEG and MEG. }
\label{tab:reconstruction}
\setlength{\tabcolsep}{6pt} % Adjust column separation
\resizebox{\textwidth}{!}{%
\begin{tabular}{llcccccccc}
\toprule
\multirow{2}{*}{\textbf{Method}}& \multirow{2}{*}{\textbf{Dataset}}
& \multicolumn{2}{c}{Low-level} & \multicolumn{5}{c}{High-level} \\
\cmidrule(r){3-4} \cmidrule(l){5-9}
 &  & PixCorr $\uparrow$ & SSIM $\uparrow$ 
        & AlexNet(2) $\uparrow$ & AlexNet(5) $\uparrow$ 
        & Inception $\uparrow$ & CLIP $\uparrow$ & SwAV $\downarrow$  \\
\midrule
\addlinespace[2pt]
\multirow{3}{*}{MEG} 
& B.D.   & 0.076 & 0.336 & 0.736 & 0.826 & 0.671 & 0.767 & \textbf{0.584} \\
& ATM  & 0.104 & \textbf{0.340} & 0.613 & 0.672 & 0.619 & 0.603 & 0.651 \\
& \textbf{Ours} & \textbf{0.137} & 0.292 & \textbf{0.737} & \textbf{0.836} & \textbf{0.721} & \textbf{0.775} & 0.600 \\
\addlinespace
\cdashline{1-9}[.4pt/2pt]
\addlinespace
\multirow{5}{*}{EEG} 
& C.C.(All)   & 0.150 & 0.347 & 0.754 & 0.623 & 0.669 & 0.715 & 0.590 \\
& C.C.(Image)  & 0.132 & 0.321 & 0.813 & 0.671 & 0.664 & 0.715 & 0.590 \\
& C.C.(Depth) & 0.104 & \textbf{0.370} & 0.796 & 0.638 & 0.565 & 0.579 & 0.686 \\
& C.C.(Text) & 0.102 & 0.288 & 0.727 & 0.582 & 0.586 & 0.598 & 0.673 \\
& \textbf{Ours}     & \textbf{0.195} & 0.336 & \textbf{0.843} & \textbf{0.905} & \textbf{0.756} & \textbf{0.808} & \textbf{0.554} \\
\addlinespace
\cdashline{1-9}[.4pt/2pt]
\addlinespace
\multirow{2}{*}{EEG (subj-8)} 
% & \textbf{Ours} (w/o VAE)  & 0.184 & 0.341 & 0.836 & 0.905 & 0.771 & 0.813 & 0.548 \\
& ATM & 0.160 & 0.345 & 0.776 & 0.866 & 0.734 & 0.786 & 0.582 \\
& \textbf{Ours} & \textbf{0.227} & \textbf{0.361} & \textbf{0.878} & \textbf{0.924} & \textbf{0.796} & \textbf{0.826} & \textbf{0.531} \\
\bottomrule
\end{tabular}%
}
\end{table}

%% file: tabs/retri_ablation.tex
\begin{table}[t]
\centering
\caption{Ablation on EEG retrieval: average top-1/top-5 accuracy (\%) for 200-way zero-shot; we compare single encoders, pairwise, and triple combinations.}
\label{tab:eeg_ablation}
\footnotesize
\setlength{\tabcolsep}{6pt}
\renewcommand{\arraystretch}{1.05}
\begin{tabular}{l l l l l l}
\toprule
\multirow{2}{*}{\textbf{Setting}} 
& \multirow{2}{*}{\textbf{Configuration}} 
& \multicolumn{2}{l}{\textbf{Intra-subject}} 
& \multicolumn{2}{l}{\textbf{Inter-subject}} \\
& & Top-1 & Top-5 & Top-1 & Top-5 \\
\midrule
\multirow{3}{*}{Individual module} 
% & H14  & 46.3 & 75.9 & 13.6 & 31.1 \\
& B32  & 52.2 & 83.3 & 13.3 & 33.9 \\
& RN50 & 48.1 & 80.4 & 12.7 & 31.7 \\
& VAE  & 44.3 & 75.2 & 10.2 & 23.9 \\
\midrule
\multirow{3}{*}{Pairwise combination}    
% & H14 + B32      & 53.4 & 82.6 & 15.2 & 34.3 \\
& RN50 + B32     & 56.9 & 86.1 & 14.4 & 36.8 \\
& RN50 + VAE     & 65.8 & 90.4 & 17.4 & 37.3 \\
& B32 + VAE      & 73.6 & 94.3 & 19.1 & 41.2 \\
\midrule
\multirow{1}{*}{Triple combination}  
% & H14 + B32 + VAE   & - & - & - & - \\
% & H14 + B32 + RN50  & - & - & - & - \\
& RN50 + B32 + VAE  & \textbf{75.7} & \textbf{94.6} & \textbf{20.0} & \textbf{44.1} \\
% & RN50 + DINO + VAE  & - & - & - & - \\
\bottomrule
\end{tabular}
\end{table}

%% file: tabs/recon_ablation.tex
% \begin{table}[h]
% \centering
% \caption{Ablation study on the effect of different fusion priors for EEG-to-Image reconstruction. We report low-level (PixCorr, SSIM) and high-level perceptual similarity metrics (AlexNet, Inception, CLIP, SwAV).}
% \label{tab:reconstruction}
% \setlength{\tabcolsep}{6pt} % Adjust column separation
% \resizebox{\textwidth}{!}{%
% \begin{tabular}{lcccccccc}
% \toprule
% & \multicolumn{2}{c}{Low-level} & \multicolumn{5}{c}{High-level} \\
% \cmidrule(r){2-3} \cmidrule(l){4-8}
% Prior Setting & PixCorr $\uparrow$ & SSIM $\uparrow$ 
%        & AlexNet(2) $\uparrow$ & AlexNet(5) $\uparrow$ 
%        & Inception $\uparrow$ & CLIP $\uparrow$ & SwAV $\downarrow$  \\
% \midrule
% \addlinespace[2pt]
% \midrule
% H14 & 0.174 & 0.327 & 0.825 & 0.872 & 0.733 & 0.773 & 0.574 \\
% % RN50 + B32   & 0.076 & 0.336 & 0.736 & 0.826 & 0.671 & 0.767 & 0.584 \\
% H14 + B32  & 0.187 & 0.340 & 0.836 & 0.908 & 0.783 & 0.814 & 0.547 \\
% H14 + VAE  & - & - & - & - & - & - & - \\
% % RN50 + B32 + VAE   & 0.150 & 0.347 & 0.754 & 0.623 & 0.669 & 0.715 & 0.590 \\
% H14 + B32 + VAE  & 0.195 & 0.336 & 0.843 & 0.905 & 0.756 & 0.808 & 0.554 \\
% \bottomrule
% \end{tabular}%
% }
% \end{table}
\begin{table}[h]
\centering
\caption{Ablation study on the effect of different fusion priors for Brain-to-Image reconstruction.}
\label{tab:ablation_reconstruction}
\setlength{\tabcolsep}{5pt}
\resizebox{\textwidth}{!}{%
\begin{tabular}{lcccccccc}
\toprule
Prior Setting
& PixCorr $\uparrow$ & SSIM $\uparrow$ 
& AlexNet(2) $\uparrow$ & AlexNet(5) $\uparrow$ 
& Inception $\uparrow$ & CLIP $\uparrow$ & SwAV $\downarrow$  \\
\midrule
H14              & 0.174  & 0.327  & 0.825  & 0.872  & 0.733  & 0.773 & 0.574  \\
H14 + B32        & 0.187  & \textbf{0.340}  & 0.836  & \textbf{0.908}  & \textbf{0.783} & \textbf{0.814} & \textbf{0.547} \\
H14 + VAE        & 0.173  & 0.312  & 0.789  & 0.838  & 0.672  & 0.721  & 0.611 \\
H14 + B32 + VAE  & \textbf{0.195}  & 0.336 & \textbf{0.843}  & 0.905  & 0.756 & 0.808  & 0.554 \\
\bottomrule
\end{tabular}%
}
\end{table}

%% file: tabs/backbone_eeg.tex
\begin{table*}[t]
  \centering
  % \small
  \setlength{\tabcolsep}{3pt}
  \caption{
  % \textcolor{BurntOrange}{Top-1 and top-5 accuracy (\%) for 200-way zero-shot retrieval on THINGS-EEG across brain encoder backbones (rows) and visual-encoder combinations (columns), averaged over all 10 subjects and trained with the same fusion-based visual interface and hyperparameters, varying only the brain backbone. \emph{Full} denotes using all three visual encoders (RN50, B32, and VAE).}
  Top-1 and top-5 accuracy (\%) for 200-way zero-shot retrieval on THINGS-EEG across brain encoder backbones (rows) and visual-encoder combinations (columns), averaged over all 10 subjects and trained with the same fusion-based visual interface and hyperparameters, varying only the brain backbone. \emph{Full} denotes using all three visual encoders (RN50, B32, and VAE).
}
  \resizebox{\textwidth}{!}{%
  \begin{tabular}{l*{7}{cc}}
    \toprule
    & \multicolumn{2}{c}{RN50}
    & \multicolumn{2}{c}{B32}
    & \multicolumn{2}{c}{VAE}
    & \multicolumn{2}{c}{RN50+B32}
    & \multicolumn{2}{c}{RN50+VAE}
    & \multicolumn{2}{c}{B32+VAE}
    & \multicolumn{2}{c}{Full} \\
    \cmidrule(lr){2-3}
    \cmidrule(lr){4-5}
    \cmidrule(lr){6-7}
    \cmidrule(lr){8-9}
    \cmidrule(lr){10-11}
    \cmidrule(lr){12-13}
    \cmidrule(lr){14-15}
    Backbone
    & top-1 & top-5
    & top-1 & top-5
    & top-1 & top-5
    & top-1 & top-5
    & top-1 & top-5
    & top-1 & top-5
    & top-1 & top-5 \\
    \midrule
    ShallowNet      & 32.4 & 64.9 & 35.7 & 69.9 & 18.5 & 43.4 & 38.2 & 72.3 & 34.4 & 66.6 & 38.7 & 72.1 & 40.8 & 73.9 \\
    DeepNet         & 17.7 & 44.9 & 18.0 & 46.9 &  6.4 & 20.7 & 19.0 & 48.7 & 11.9 & 36.1 & 17.3 & 43.9 & 17.8 & 45.1 \\
    EEGNet          & 31.5 & 62.0 & 34.6 & 67.3 & 23.0 & 50.3 & 37.6 & 70.8 & 39.1 & 70.0 & 43.7 & 75.7 & 45.4 & 77.1 \\
    TSConv          & 38.7 & 72.0 & 42.2 & 74.5 & 25.8 & 55.6 & 45.5 & 78.1 & 46.4 & 78.1 & 52.5 & 83.0 & 54.5 & 84.1 \\
    ATM             & 41.5 & 74.1 & 42.9 & 76.0 & 34.4 & 63.9 & 46.8 & 78.2 & 52.2 & 81.4 & 56.2 & 85.4 & 57.9 & 86.5 \\
    BrainProjection & 47.1 & 78.7 & 50.5 & 82.4 & 43.0 & 74.6 & 55.4 & 85.0 & 64.0 & 89.8 & 72.1 & 93.0 & 72.9 & 94.1 \\
    MBP (ours) & \textbf{48.1} & \textbf{80.5} & \textbf{51.2} & \textbf{83.0} & \textbf{44.1} & \textbf{75.5} & \textbf{56.2} & \textbf{86.2} & \textbf{65.2} & \textbf{90.7} & \textbf{73.7} & \textbf{94.1} & \textbf{75.7} & \textbf{94.6} \\
    \bottomrule
  \end{tabular}%
  }

  \label{tab:backbone_eeg}
\end{table*}

%% file: sec/5_conclusion.tex
\section{Conclusion}
% \textcolor{ForestGreen}{In this paper, we study learning brain representation with hierarchical visual embeddings for brain-to-image decoding. We propose a fusion-based brain–vision interface that aligns brain signals to a single token built from complementary semantic and pixel-level encoders and feeds it into a frozen generation prior. Experiments on THINGS-EEG and THINGS-MEG show that this interface achieves strong 200-way zero-shot retrieval in both intra- and inter-subject settings, together with high-quality reconstructions in intra-subject decoding. Ablation studies show that fusing CLIP and VAE features improves brain-to-image decoding performance over single-encoder and semantics-only baselines. Plugging various EEG backbones into the same interface yields consistent retrieval gains without retraining the visual side, highlighting hierarchical visual embeddings as a plug-and-play route to robust brain representations for brain-to-visual decoding.}
In this paper, we study learning brain representation with hierarchical visual embeddings for brain-to-image decoding. We propose a fusion-based brain–vision interface that aligns brain signals to a single token built from complementary semantic and pixel-level encoders and feeds it into a frozen generation prior. Experiments on THINGS-EEG and THINGS-MEG show that this interface achieves strong 200-way zero-shot retrieval in both intra- and inter-subject settings, together with high-quality reconstructions in intra-subject decoding. Ablation studies show that fusing CLIP and VAE features improves brain-to-image decoding performance over single-encoder and semantics-only baselines. Plugging various EEG backbones into the same interface yields consistent retrieval gains without retraining the visual side, highlighting hierarchical visual embeddings as a plug-and-play route to robust brain representations for brain-to-visual decoding.
% \textcolor{BrickRed}{\paragraph{Limitations and future work}
% Our experiments are restricted to THINGS-EEG and THINGS-MEG; extending the proposed interface to other datasets, modalities, and tasks is an important direction for future work. The current fusion design and visual stack represent a single hand-picked configuration, and exploring stronger vision backbones and alternative encoder sets may further improve decoding performance and efficiency.}
\paragraph{Limitations and future work}
Our experiments are restricted to THINGS-EEG and THINGS-MEG; extending the proposed interface to other datasets, modalities, and tasks is an important direction for future work. The current fusion design and visual stack represent a single hand-picked configuration, and exploring stronger vision backbones and alternative encoder sets may further improve decoding performance and efficiency.

\section*{Acknowledgments}
This work was supported by the National Natural Science Foundation of China under Grant No. 62502411.

%% file: sec/6_appendix.tex
\clearpage
\appendix
\section{LLM Usage Statement}
We used LLM for grammar checking and language polishing to improve readability.

\section{Datasets Details}
\label{appendix:dataset}
\paragraph{THINGS-EEG} THINGS-EEG~\citep{alessandro2022large} is a large-scale dataset of electroencephalography (EEG) recordings from 10 participants. Signals are acquired with a 64-channel EASYCAP arranged according to the international 10–10 system. The training split spans 1,654 object concepts, each represented by 10 images; every image is shown four times to each participant (1,654 × 10 × 4). The test split covers 200 concepts with a single image per concept, repeated 80 times (200 × 1 × 80). Preprocessing follows \citet{song2024decoding,wu2025bridging}: raw EEG is filtered to 0.1–100 Hz, yielding 63 channels at 1,000 Hz; trials are segmented from 0–1,000 ms post-stimulus with baseline correction using the prior 200 ms average. Data is then downsampled to 250 Hz, and 17 posterior channels over occipital and parietal sites associated with visual processing are retained. To improve signal-to-noise ratio, repetitions are averaged, producing 16,540 training samples and 200 test samples per participant.

\paragraph{THINGS-MEG} THINGS-MEG \citep{hebart2023things} is a large-scale dataset of magnetoencephalography (MEG) recordings from 4 participants. Signals are acquired with 271 channels. Each trial presents an image for 500 ms, followed by a blank screen of 1000 ± 200 ms. The training split spans 1,854 object concepts, each represented by 12 images; every image is shown once to each participant (1,854 × 12 × 1). The test split covers 200 concepts with a single image per concept, repeated 12 times (200 × 1 × 12). To construct the zero-shot task, 200 test concepts are discarded from the training set. Preprocessing follows \citet{song2024decoding,wu2025bridging}: raw MEG is filtered to 0.1–100 Hz; trials are segmented from 0–1,000 ms post-stimulus with baseline correction. Data is then downsampled to 200 Hz. To improve signal-to-noise ratio, repetitions are averaged, producing 19,848 training samples and 200 test samples per participant.
% \textcolor{BurntOrange}{
% \paragraph{Ethical considerations} Our experiments rely exclusively on the publicly released THINGS-EEG/MEG datasets~\citep{alessandro2022large,hebart2023things} collected under informed consent and institutional oversight, and we use only anonymized recordings. While brain-to-image decoding may ultimately benefit populations such as locked-in or speech-impaired patients, it also poses clear risks, including unauthorized inference of mental states and privacy violations. We view our method as a technical contribution on a benchmark dataset, not as a tool for covert monitoring, and we stress that any real-world deployment must require explicit consent, robust data protection, and adherence to clinical and legal regulations.
% }
\paragraph{Ethical considerations} Our experiments rely exclusively on the publicly released THINGS-EEG/MEG datasets~\citep{alessandro2022large,hebart2023things} collected under informed consent and institutional oversight, and we use only anonymized recordings. While brain-to-image decoding may ultimately benefit populations such as locked-in or speech-impaired patients, it also poses clear risks, including unauthorized inference of mental states and privacy violations. We view our method as a technical contribution on a benchmark dataset, not as a tool for covert monitoring, and we stress that any real-world deployment must require explicit consent, robust data protection, and adherence to clinical and legal regulations.

\section{Results Details}
\paragraph{Per-Subject retrieval on THINGS-EEG and THINGS-MEG}
We report 200-way zero-shot Top-1/Top-5 accuracy per subject for THINGS-EEG and THINGS-MEG. 
For each subject, we evaluate individual encoders (RN50, B32, VAE), pairwise stacks (RN50{+}B32, RN50{+}VAE, B32{+}VAE), and the triple stack (RN50{+}B32{+}VAE) with the VAE input fixed at $128{\times}128$.
% Subject-wise means are listed in the last row; intra- and inter-subject protocols follow the same evaluation as described in the main text.

\input{tabs/appendix_retri_eeg_baseline}
\input{tabs/appendix_eeg_retri}

\input{tabs/appendix_retri_meg_baseline}
\input{tabs/appendix_meg_retri}

\input{tabs/appendix_recon_eeg_h14}
\input{tabs/appendix_recon_eeg_h14+b32}
\input{tabs/appendix_recon_eeg_h14+vae}
\input{tabs/appendix_recon_eeg_h14+b32+vae}
\paragraph{Per-Subject reconstruction metrics}
We further report reconstruction metrics per subject in Tab.~\ref{tab:eeg_recon_h14} Tab.~\ref{tab:eeg_recon_h14+b32}, Tab.~\ref{tab:eeg_recon_h14+vae}, and Tab.~\ref{tab:eeg_recon_h14+b32+vae}. For each subject, we compute low-level measures (PixCorr, SSIM) and high-level perceptual similarity (AlexNet(2/5), Inception, CLIP) with SwAV↓ as a diversity/consistency proxy. Results are shown for the single target (H14 only), semantic pair (H14{+}B32, H14{+}VAE) and the full multiscale stack (H14{+}B32{+}VAE). The last row gives subject-wise means.

\begin{figure}[h]
    \centering
    \includegraphics[width=\linewidth]{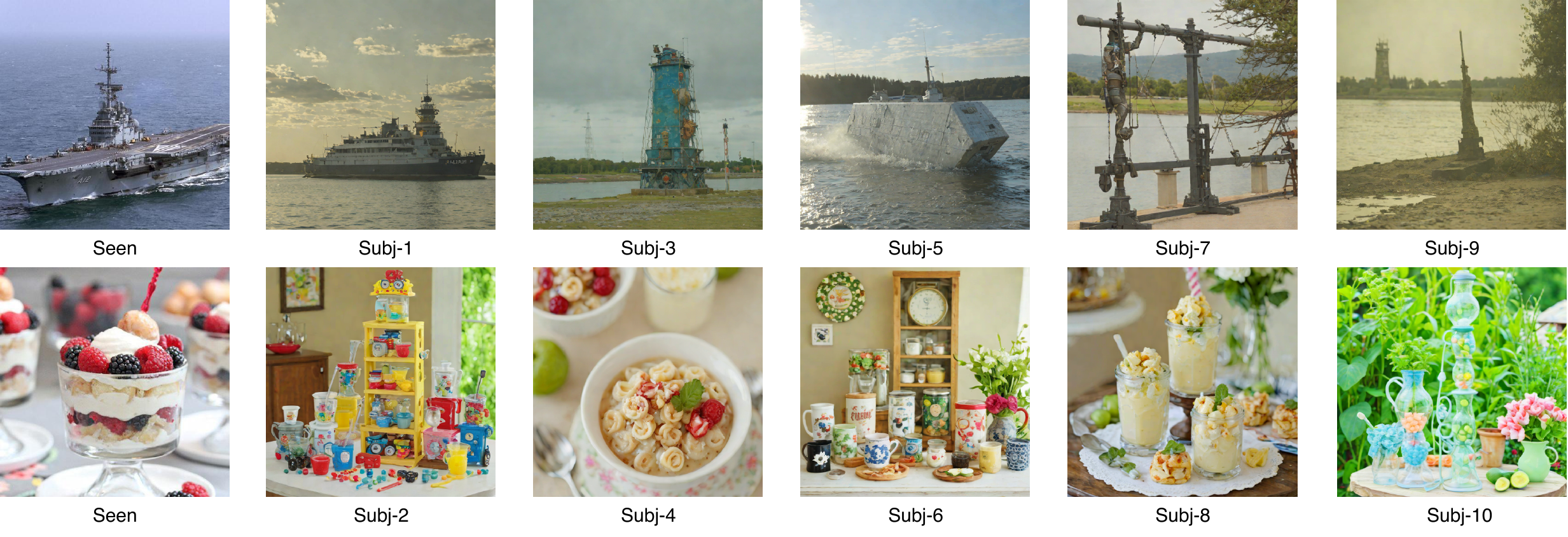}
    \caption{\textbf{Cross-subject EEG reconstructions.}}
    \label{fig:cross-subjection}
\end{figure}

\paragraph{Reconstruction from different subjects}
As shown in Fig.~\ref{fig:cross-subjection}, for the same visual stimulus, we reconstruct images from EEG recorded from different subjects.

\clearpage

\section{Additional Ablation Studies}
\begin{figure}[t]
  \centering
  \includegraphics[width=\linewidth]{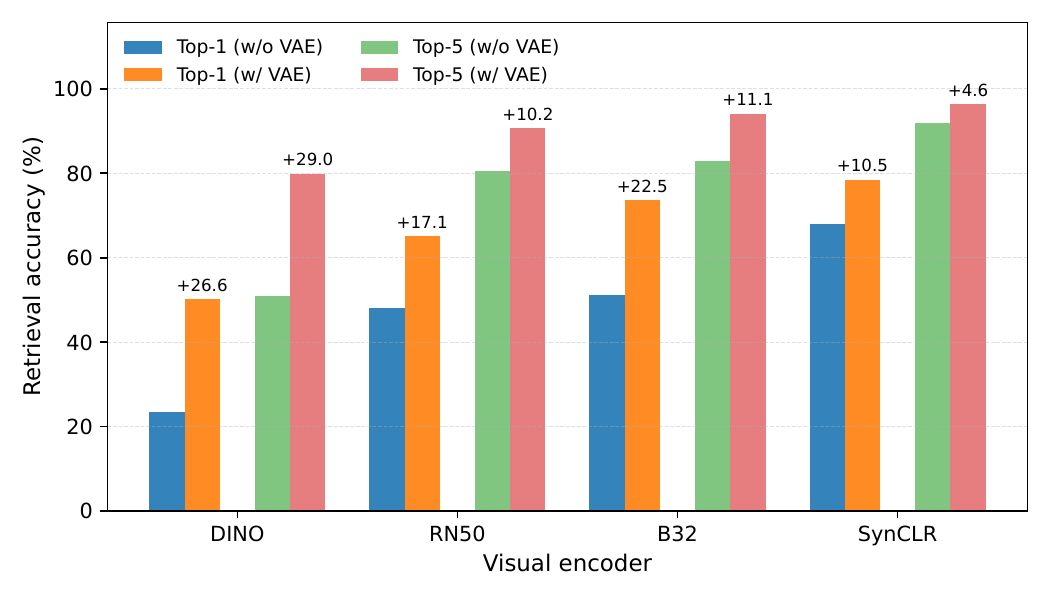}
  \caption{
  % \textcolor{BurntOrange}{Top-1 and top-5 accuracy (\%) for 200-way zero-shot retrieval on THINGS-EEG when aligning EEG to single encoders versus encoder+VAE mixtures. Bars show per-setting performance and absolute gains from adding the SDXL VAE, averaged over all 10 subjects.}
  Top-1 and top-5 accuracy (\%) for 200-way zero-shot retrieval on THINGS-EEG when aligning EEG to single encoders versus encoder+VAE mixtures. Bars show per-setting performance and absolute gains from adding the SDXL VAE, averaged over all 10 subjects.
  }
  \label{fig:retrieval_single_vs_mixed_encoder_bar}
\end{figure}

% \textcolor{BurntOrange}{
% \paragraph{Extended visual encoder ablations}
% Fig.~\ref{fig:retrieval_single_vs_mixed_encoder_bar} and Tab.~\ref{tab:visual_ablation_extended} compare additional visual encoders (DINO~\citep{oquab2023dinov2} and SynCLR~\citep{tian2024learning}) with their multi-encoder variants. Across backbones, adding the SDXL VAE consistently boosts top-1 and top-5 retrieval accuracy in both intra- and inter-subject settings, with especially large gains for DINO. Pairing DINO with CLIP encoders (RN50 or B32) brings moderate improvements, while VAE-based combinations yield the strongest gains, particularly when fused with B32. For SynCLR, stacking with RN50 or B32 alone largely saturates performance and can slightly hurt inter-subject transfer, whereas SynCLR+VAE and VAE-based triple stacks achieve the best overall results. These trends indicate that low-level VAE features systematically complement high-level representations for EEG--image alignment across visual backbones.
% }
\paragraph{Extended visual encoder ablations}
Fig.~\ref{fig:retrieval_single_vs_mixed_encoder_bar} and Tab.~\ref{tab:visual_ablation_extended} compare additional visual encoders (DINO~\citep{oquab2023dinov2} and SynCLR~\citep{tian2024learning}) with their multi-encoder variants. Across backbones, adding the SDXL VAE consistently boosts top-1 and top-5 retrieval accuracy in both intra- and inter-subject settings, with especially large gains for DINO. Pairing DINO with CLIP encoders (RN50 or B32) brings moderate improvements, while VAE-based combinations yield the strongest gains, particularly when fused with B32. For SynCLR, stacking with RN50 or B32 alone largely saturates performance and can slightly hurt inter-subject transfer, whereas SynCLR+VAE and VAE-based triple stacks achieve the best overall results. These trends indicate that low-level VAE features systematically complement high-level representations for EEG--image alignment across visual backbones.
\begin{figure}[t]
\centering
\includegraphics[width=0.8\linewidth]{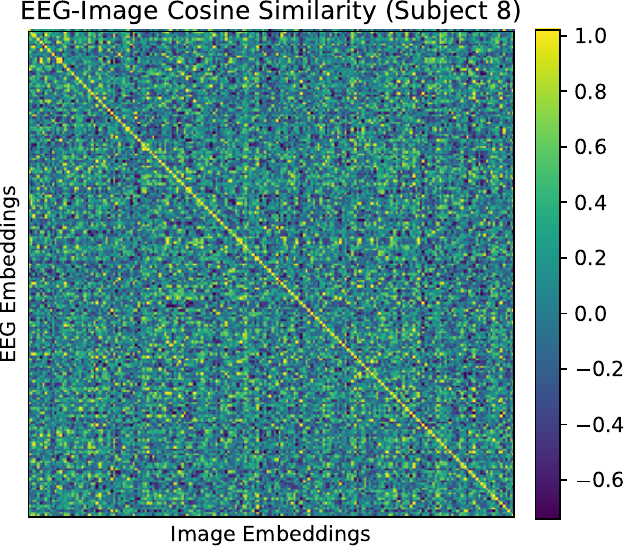}
\caption{
% \textcolor{Purple}{Cosine similarity matrix between EEG and image embeddings for Subject~8 on the test split of the THINGS-EEG dataset.
% Rows correspond to EEG embeddings and columns to image embeddings; the strong diagonal pattern indicates that each EEG embedding is most similar to its paired image embedding.}
Cosine similarity matrix between EEG and image embeddings for Subject~8 on the test split of the THINGS-EEG dataset.
Rows correspond to EEG embeddings and columns to image embeddings; the strong diagonal pattern indicates that each EEG embedding is most similar to its paired image embedding.
}
\label{fig:similarity_eeg_image_subj8}
\end{figure}
\input{tabs/appendix_extended_visual}
% \textcolor{BurntOrange}{
% \paragraph{Effect of EEG feature selection on channels}
% Unless otherwise stated, we use EEG data from 0–1000~ms after stimulus onset and restrict inputs to 17 occipital–parietal (O+P\footnote{P7,P5,P3,P1,Pz,P2,P4,P6,P8,PO7,PO3,POz,PO4,PO8,O1,Oz,O2}) channels following previous work~\citep{song2024decoding}. Tab.~\ref{tab:channel_ablation} shows that, for each visual-encoder configuration, the O+P subset consistently attains the highest 200-way retrieval accuracy compared to using only occipital channels, only parietal channels, ''other'' non-visual channels, or the full 63-channel montage.
% }
\paragraph{Effect of EEG feature selection on channels}
Unless otherwise stated, we use EEG data from 0–1000~ms after stimulus onset and restrict inputs to 17 occipital–parietal (O+P\footnote{P7,P5,P3,P1,Pz,P2,P4,P6,P8,PO7,PO3,POz,PO4,PO8,O1,Oz,O2}) channels following previous work~\citep{song2024decoding}. Tab.~\ref{tab:channel_ablation} shows that, for each visual-encoder configuration, the O+P subset consistently attains the highest 200-way retrieval accuracy compared to using only occipital channels, only parietal channels, ''other'' non-visual channels, or the full 63-channel montage.
\input{tabs/appendix_channels}
\begin{figure}[t]
  \centering
  \includegraphics[width=\linewidth]{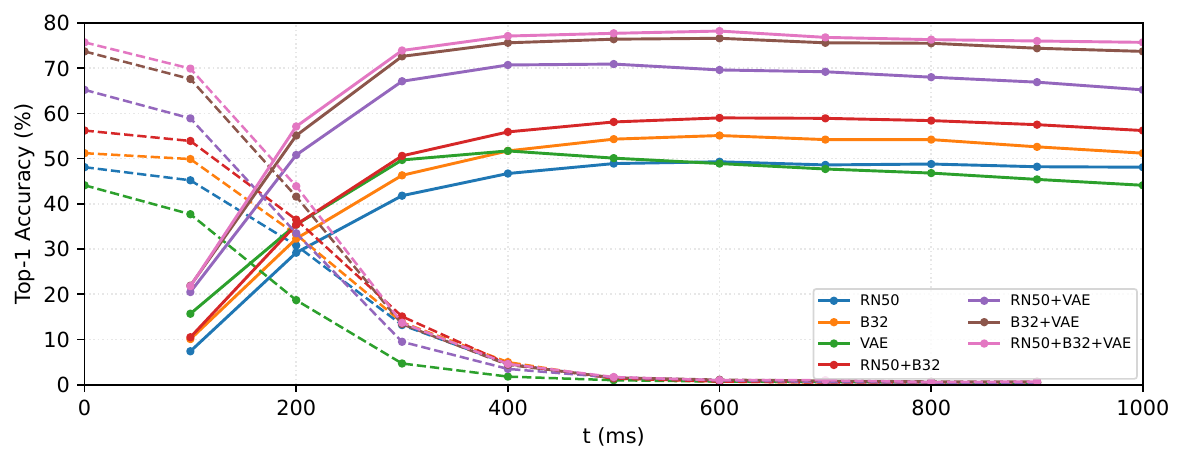}
  \caption{
    % \textcolor{Purple}{Top-1 retrieval accuracy (\%) as a function of the EEG time window for different visual encoder configurations.
    % Solid lines show cumulative windows $[0, t]$; dashed lines show tail windows $[t, 1000]$, with $t{=}0$ denoting the full $[0, 1000]$~ms interval.}
    Top-1 retrieval accuracy (\%) as a function of the EEG time window for different visual encoder configurations.
    Solid lines show cumulative windows $[0, t]$; dashed lines show tail windows $[t, 1000]$, with $t{=}0$ denoting the full $[0, 1000]$~ms interval.
  }
  \label{fig:time_windows_combined}
\end{figure}
% \textcolor{Purple}{
% \paragraph{Effect of EEG feature selection on time windows}
% Fig.~\ref{fig:time_windows_combined} plots top-1 retrieval accuracy as a function of the EEG time window for different visual encoder configurations. For cumulative windows $[0,t]$, accuracy under the full RN50+B32+VAE fusion setting rises from 21.8\% at 0–100~ms to 73.9\% at 0–300~ms and then stays close to the 0–1000~ms baseline (75.7\%), while tail windows $[t,1000]$ that exclude early activity drop from 75.7\% (0–1000~ms) to 69.9\% (100–1000~ms) and 13.7\% (300–1000~ms), approaching chance for 500–1000~ms (1.7\%). Comparing encoders, VAE-based configurations tend to outperform CLIP-based ones in the earliest cumulative windows, whereas CLIP-based configurations outperform VAE in later tail windows, suggesting that early EEG activity may be more aligned with low-level, pixel-like features and later activity may be more aligned with higher-level semantic features. For more complete windows (i.e., larger t for [0, t] and smaller t for [t, 1000]), CLIP-dominated and fused configurations generally surpass VAE alone. This is consistent with the possibility that, after a few hundred milliseconds, the EEG signals most useful for 200-way zero-shot retrieval are dominated by hig1her-level visual information. Across all time windows, the full RN50+B32+VAE fusion consistently achieves the best performance, further supporting the effectiveness of our approach.
% }
\paragraph{Effect of EEG feature selection on time windows}
Fig.~\ref{fig:time_windows_combined} plots top-1 retrieval accuracy as a function of the EEG time window for different visual encoder configurations. For cumulative windows $[0,t]$, accuracy under the full RN50+B32+VAE fusion setting rises from 21.8\% at 0–100~ms to 73.9\% at 0–300~ms and then stays close to the 0–1000~ms baseline (75.7\%), while tail windows $[t,1000]$ that exclude early activity drop from 75.7\% (0–1000~ms) to 69.9\% (100–1000~ms) and 13.7\% (300–1000~ms), approaching chance for 500–1000~ms (1.7\%). Comparing encoders, VAE-based configurations tend to outperform CLIP-based ones in the earliest cumulative windows, whereas CLIP-based configurations outperform VAE in later tail windows, suggesting that early EEG activity may be more aligned with low-level, pixel-like features and later activity may be more aligned with higher-level semantic features. For more complete windows (i.e., larger t for [0, t] and smaller t for [t, 1000]), CLIP-dominated and fused configurations generally surpass VAE alone. This is consistent with the possibility that, after a few hundred milliseconds, the EEG signals most useful for 200-way zero-shot retrieval are dominated by hig1her-level visual information. Across all time windows, the full RN50+B32+VAE fusion consistently achieves the best performance, further supporting the effectiveness of our approach.
\input{tabs/appendix_mask}
% \textcolor{BurntOrange}{\paragraph{Encoder-masking analysis}
% To probe the relative contribution of each visual branch in the RN50+B32+VAE fusion model, we keep the trained model fixed and zero-mask individual encoder embeddings only at inference time (Tab.~\ref{tab:encoder_masking}). Removing the VAE or B32 branches causes substantial drops in both intra- and inter-subject accuracy, whereas masking RN50 has a comparatively smaller effect, indicating that VAE and B32 provide the dominant complementary signals in this fusion configuration.}
\paragraph{Encoder-masking analysis}
To probe the relative contribution of each visual branch in the RN50+B32+VAE fusion model, we keep the trained model fixed and zero-mask individual encoder embeddings only at inference time (Tab.~\ref{tab:encoder_masking}). Removing the VAE or B32 branches causes substantial drops in both intra- and inter-subject accuracy, whereas masking RN50 has a comparatively smaller effect, indicating that VAE and B32 provide the dominant complementary signals in this fusion configuration.

%% file: tabs/appendix_retri_eeg_baseline.tex
\begin{table*}[h]
\centering
\caption{Top-1 and Top-5 accuracy (\%) for 200-way zero-shot retrieval on THINGS-EEG.}
\label{tab:things-eeg}
\resizebox{\textwidth}{!}{%
\begin{tabular}{l*{11}{cc}}
\toprule
\multirow{2}{*}{\vspace{-0.38em}\textbf{Method}}
& \multicolumn{2}{c}{\textbf{Sub1}}
& \multicolumn{2}{c}{\textbf{Sub2}}
& \multicolumn{2}{c}{\textbf{Sub3}}
& \multicolumn{2}{c}{\textbf{Sub4}}
& \multicolumn{2}{c}{\textbf{Sub5}}
& \multicolumn{2}{c}{\textbf{Sub6}}
& \multicolumn{2}{c}{\textbf{Sub7}}
& \multicolumn{2}{c}{\textbf{Sub8}}
& \multicolumn{2}{c}{\textbf{Sub9}}
& \multicolumn{2}{c}{\textbf{Sub10}}
& \multicolumn{2}{c}{\textbf{Avg}} \\
\cmidrule(lr){2-3}\cmidrule(lr){4-5}\cmidrule(lr){6-7}\cmidrule(lr){8-9}\cmidrule(lr){10-11}
\cmidrule(lr){12-13}\cmidrule(lr){14-15}\cmidrule(lr){16-17}\cmidrule(lr){18-19}\cmidrule(lr){20-21}\cmidrule(lr){22-23}
& top-1 & top-5 & top-1 & top-5 & top-1 & top-5 & top-1 & top-5 & top-1 & top-5
& top-1 & top-5 & top-1 & top-5 & top-1 & top-5 & top-1 & top-5 & top-1 & top-5 & top-1 & top-5 \\
\midrule
% \multicolumn{23}{c}{\textit{Intra-subject: train and test on one subject}} \\
% \midrule
BraVL
& 6.1 & 17.9 & 4.9 & 14.9 & 5.6 & 17.4 & 5.0 & 15.1 & 4.0 & 13.4
& 6.0 & 18.2 & 6.5 & 20.4 & 8.8 & 23.7 & 4.3 & 14.0 & 7.0 & 19.7 & 5.8 & 17.5 \\
NICE
& 13.2 & 39.5 & 13.5 & 40.3 & 14.5 & 42.7 & 20.6 & 52.7 & 10.1 & 31.5
& 16.5 & 44.0 & 17.0 & 42.1 & 22.9 & 56.1 & 15.4 & 41.6 & 17.4 & 45.8 & 16.1 & 43.6 \\
NICE-SA
& 13.3 & 40.2 & 12.1 & 36.1 & 15.3 & 39.6 & 15.9 & 49.0 & 9.8 & 34.4
& 14.2 & 42.4 & 17.9 & 43.6 & 18.2 & 50.2 & 14.4 & 38.7 & 16.0 & 42.8 & 14.7 & 41.7 \\
NICE-GA
& 15.2 & 40.1 & 13.9 & 40.1 & 14.7 & 42.7 & 17.6 & 48.9 & 9.0 & 29.7
& 16.4 & 44.4 & 14.9 & 43.1 & 20.3 & 52.1 & 14.1 & 39.7 & 19.6 & 46.7 & 15.6 & 42.8 \\
MB2C
& 23.7 & 56.3 & 22.7 & 50.5 & 26.3 & 60.2 & 34.8 & 67.0 & 21.3 & 53.0
& 31.0 & 62.3 & 25.0 & 54.8 & 39.0 & 69.3 & 27.5 & 59.3 & 33.2 & 70.8 & 28.5 & 60.4 \\
ATM-S
& 25.6 & 60.4 & 22.0 & 54.5 & 25.0 & 62.4 & 31.4 & 60.9 & 12.9 & 43.0
& 21.3 & 51.1 & 30.5 & 61.5 & 38.8 & 72.0 & 34.4 & 51.5 & 29.1 & 63.5 & 28.5 & 60.4 \\
VE-SDN
& 32.6 & 63.7 & 34.4 & 69.9 & 38.7 & 73.5 & 39.8 & 72.0 & 29.4 & 58.6
& 34.5 & 68.8 & 34.5 & 68.3 & 49.3 & 79.8 & 39.0 & 69.6 & 39.8 & 75.3 & 37.2 & 69.9 \\
CognitionCapturer-All
& 31.4 & 79.7 & 31.4 & 77.8 & 38.2 & 85.7 & 40.4 & 85.8 & 24.4 & 66.3
& 34.8 & 78.8 & 34.7 & 81.0 & 48.1 & 88.6 & 31.4 & 79.4 & 35.6 & 79.3 & 35.6 & 80.2 \\
UBP
& 41.2 & 70.5 & 51.2 & 80.9 & 51.2 & 82.0 & 51.1 & 76.9 & 42.2 & 72.8
& 57.5 & 83.5 & 49.0 & 79.9 & 58.6 & 85.8 & 45.1 & 76.2 & 61.5 & 88.2 & 50.9 & 79.7 \\
Ours
& 64.3 & 88.8 & 76.3 & 95.3 & 74.0 & 95.0 & 67 & 91.8 & 68.0 & 91.5
& 81.5 & 96.3 & 76.8 & 96.8 & 84.8 & 98.5 & 76.8 & 95.8 & 87.3 & 99.3 & 75.7 & 94.6 \\
% \midrule
% \multicolumn{23}{c}{\textit{Inter-subject: leave one Subout for test}} \\
% \midrule
% BraVL
% & 2.3 & 8.0 & 1.5 & 6.3 & 1.4 & 5.9 & 1.7 & 6.7 & 1.5 & 5.6
% & 1.8 & 7.2 & 2.1 & 8.1 & 2.2 & 7.6 & 1.6 & 6.4 & 2.3 & 8.5 & 1.8 & 7.0 \\
% NICE
% & 7.6 & 22.8 & 5.9 & 20.5 & 6.0 & 22.3 & 6.3 & 20.7 & 4.4 & 18.3
% & 5.6 & 22.2 & 5.6 & 19.7 & 6.3 & 22.0 & 5.7 & 17.6 & 8.4 & 28.3 & 6.2 & 21.4 \\
% NICE-SA
% & 7.0 & 22.6 & 6.6 & 23.2 & 7.5 & 23.7 & 5.4 & 21.4 & 6.4 & 22.2
% & 7.5 & 22.5 & 3.8 & 19.1 & 8.5 & 24.4 & 7.4 & 22.3 & 9.8 & 29.6 & 7.0 & 23.1 \\
% NICE-GA
% & 5.9 & 21.4 & 6.4 & 22.7 & 5.5 & 20.1 & 6.1 & 21.0 & 4.7 & 19.5
% & 6.2 & 22.5 & 5.9 & 19.1 & 7.3 & 25.3 & 4.8 & 18.3 & 6.2 & 26.3 & 5.9 & 21.6 \\
% ATM-S
% & 10.5 & 26.8 & 7.1 & 24.8 & 11.9 & 33.8 & 14.7 & 39.4 & 7.0 & 23.9
% & 11.1 & 35.8 & 16.1 & 43.5 & 15.0 & 40.3 & 4.9 & 22.7 & 20.5 & 46.5 & 11.8 & 33.7 \\
% UBP
% & 11.5 & 29.7 & 15.5 & 40.0 & 9.8 & 27.0 & 13.0 & 32.3 & 8.8 & 33.8
% & 11.7 & 31.0 & 10.2 & 23.8 & 12.2 & 32.2 & 15.5 & 40.5 & 16.0 & 43.5 & 12.4 & 33.4 \\
% Ours
% & 27.0 & 51.5 & 27.5 & 56.5 & 10.0 & 34.0 & 22.5 & 49.0 & 22.0 & 51.5 
% & 18.5 & 44.0 & 21.5 & 48.5 & 21.5 & 44.0 & 27.0 & 60.5 & 33.5 & 66.0 & 23.1 & 50.6\\
\bottomrule
\end{tabular}}
\end{table*}

%% file: tabs/appendix_eeg_retri.tex
\begin{table*}[h]
\centering
\caption{Top-1 and Top-5 accuracy (\%) for 200-way zero-shot retrieval on THINGS-EEG across different configurations.}
% \caption{Top-1 and Top-5 accuracy (\%) for 200-way zero-shot retrieval on THINGS-EEG across different configurations.}
\label{tab:things-eeg-config}
\resizebox{\textwidth}{!}{%
\begin{tabular}{l*{11}{cc}} % 第一列 l 左对齐
\toprule
\multirow{2}{*}{\vspace{-0.38em}\textbf{Configuration}}
& \multicolumn{2}{c}{\textbf{Sub1}}
& \multicolumn{2}{c}{\textbf{Sub2}}
& \multicolumn{2}{c}{\textbf{Sub3}}
& \multicolumn{2}{c}{\textbf{Sub4}}
& \multicolumn{2}{c}{\textbf{Sub5}}
& \multicolumn{2}{c}{\textbf{Sub6}}
& \multicolumn{2}{c}{\textbf{Sub7}}
& \multicolumn{2}{c}{\textbf{Sub8}}
& \multicolumn{2}{c}{\textbf{Sub9}}
& \multicolumn{2}{c}{\textbf{Sub10}}
& \multicolumn{2}{c}{\textbf{Avg}} \\
\cmidrule(lr){2-3}\cmidrule(lr){4-5}\cmidrule(lr){6-7}\cmidrule(lr){8-9}\cmidrule(lr){10-11}
\cmidrule(lr){12-13}\cmidrule(lr){14-15}\cmidrule(lr){16-17}\cmidrule(lr){18-19}\cmidrule(lr){20-21}\cmidrule(lr){22-23}
& top-1 & top-5 & top-1 & top-5 & top-1 & top-5 & top-1 & top-5 & top-1 & top-5
& top-1 & top-5 & top-1 & top-5 & top-1 & top-5 & top-1 & top-5 & top-1 & top-5 & top-1 & top-5 \\
\midrule
B32
 & 39.3 & 75.3 & 48.8 & 79.3 & 53.3 & 84.5 & 54.8 & 87.0 & 42.8 & 75.0 & 57.8 & 84.3 & 47.0 & 81.0 & 62.3 & 89.3 & 44.3 & 80.0 & 61.8 & 94.3 & 51.2 & 83.0 \\
RN50
 & 40.0 & 69.5 & 48.5 & 79.5 & 48.5 & 85.0 & 45.8 & 82.0 & 41.5 & 74.0 & 55.8 & 83.0 & 48.5 & 77.5 & 55.5 & 88.0 & 41.8 & 77.0 & 55.3 & 89.5 & 48.1 & 80.5 \\
VAE
& 38.8 & 71.5 & 41.3 & 72.5 & 43.3 & 75.3 & 33.5 & 65.3 & 39.8 & 70.5
& 50.5 & 81.8 & 44.3 & 75.0 & 55.0 & 86.3 & 42.8 & 73.3 & 52.0 & 83.5 & 44.1 & 75.5 \\
\midrule
RN50+B32
& 47.3 & 79.0 & 55.8 & 81.0 & 56.3 & 87.5 & 59.8 & 88.8 & 46.8 & 81.0
& 63.0 & 87.5 & 53.0 & 85.0 & 65.0 & 91.3 & 50.0 & 86.5 & 68.3 & 94.5 & 56.5 & 86.2 \\
RN50+VAE
& 60.8 & 86.0 & 62.8 & 92.0 & 59.8 & 91.0 & 53.3 & 87.0 & 58.0 & 84.0
& 73.0 & 94.0 & 62.5 & 88.8 & 77.0 & 97.3 & 68.3 & 90.8 & 77.0 & 96.5 & 65.2 & 90.7 \\
B32+VAE
& 63.0 & 88.3 & 70.3 & 94.0 & 73.5 & 94.3 & 64.3 & 92.3 & 70.5 & 91.0
& 78.0 & 96.0 & 73.3 & 93.3 & 84.8 & 97.5 & 75.0 & 96.0 & 84.8 & 98.8 & 73.7 & 94.1 \\
\midrule
RN50+B32+VAE
& 64.3 & 88.8 & 76.3 & 95.3 & 74.0 & 95.0 & 67 & 91.8 & 68.0 & 91.5
& 81.5 & 96.3 & 76.8 & 96.8 & 84.8 & 98.5 & 76.8 & 95.8 & 87.3 & 99.3 & 75.7 & 94.6 \\
\bottomrule
\end{tabular}}
\end{table*}

%% file: tabs/appendix_retri_meg_baseline.tex
\begin{table}[h]
  \centering
  \caption{Top-1 and Top-5 accuracy (\%) for 200-way zero-shot retrieval on THINGS-MEG}
  \label{tab:compare_meg}
  \huge
  \resizebox{\linewidth}{!}{
  \begin{tabular}{lcccccccccc}
    \toprule
    \multirow{2}{*}{\vspace{-0.3em}\textbf{Method}}
    & \multicolumn{2}{c}{Sub1} & \multicolumn{2}{c}{Sub2} & \multicolumn{2}{c}{Sub3} & \multicolumn{2}{c}{Sub4}& \multicolumn{2}{c}{Avg} \\
    \cmidrule(r){2-3} \cmidrule(r){4-5} \cmidrule(r){6-7} \cmidrule(r){8-9} \cmidrule(r){10-11}
    & top-1 & top-5 & top-1 & top-5 & top-1 & top-5 & top-1 & top-5 & top-1 & top-5 \\
    \midrule
    % \multicolumn{11}{c}{\textbf{Intra-subject}: train and test on one subject} \\
    % \midrule
    NICE & 9.6 & 27.8 & 18.5 & 47.8 & 14.2 & 41.6 & 9.0 & 26.6 & 12.8 & 36.0\\   
    NICE-SA & 9.8 & 27.8 & 18.6 & 46.4 & 10.5 & 38.4 & 11.7 & 27.2 & 12.7 & 35.0\\ 
    NICE-GA & 8.7 & 30.5 & 21.8 & 56.6 & 16.5 & 49.7 & 10.3 & 32.3 & 14.3 & 42.3\\ 
    UBP &15.0 & 38.0 & 46.0 & 80.5 & 27.3 & 59.0 & 18.5 & 43.5 & 26.7 & 55.2 \\
    % \textbf{Ours} &17.0 & 41.0 & 60.5 & 91.0 & 42.0 & 79.0 & 22.0 & 44.5 & 35.4 & 63.9 \\
    Ours & 14.0 & 31.8 & 63.8 & 91.8 & 41.0 & 78.3 & 17.0 & 41.0 & 33.9 & 60.7 \\
    % \midrule
    % \multicolumn{11}{c}{\textbf{Inter-subject}: leave one Subout for test} \\
    % \midrule
    %  UBP &2.0 & 5.7 & 1.5 & 17.2 & 2.7 & 10.5 & 2.5 & 8.0 & 2.2 & 10.4\\ 
    %  % \textbf{Ours} &5.5 & 11.5 & 5.5 & 18.0 & 4.0 & 18.0 & 4.5 & 14.0 & 4.9 & 15.4\\
    %  Ours & 6.0 & 12.0 & 6.5 & 20.0 & 4.5 & 19.5 & 7.0 & 15.5 & 6.0 & 16.8 \\
    \bottomrule
  \end{tabular}}
\end{table}

%% file: tabs/appendix_meg_retri.tex
\begin{table*}[t]
\centering
\caption{Top-1 and Top-5 accuracy (\%) for 200-way zero-shot retrieval on THINGS-MEG across different configurations.}
\label{tab:things-eeg-config-4}
\resizebox{\textwidth}{!}{%
\begin{tabular}{l*{5}{cc}} % 4个subject + avg
\toprule
\multirow{2}{*}{\vspace{-0.38em}\textbf{Configuration}} % 竖直居中，跨两行
& \multicolumn{2}{c}{\textbf{Sub1}}
& \multicolumn{2}{c}{\textbf{Sub2}}
& \multicolumn{2}{c}{\textbf{Sub3}}
& \multicolumn{2}{c}{\textbf{Sub4}}
& \multicolumn{2}{c}{\textbf{Avg}} \\
\cmidrule(lr){2-3}\cmidrule(lr){4-5}\cmidrule(lr){6-7}\cmidrule(lr){8-9}\cmidrule(lr){10-11}
& top-1 & top-5 & top-1 & top-5 & top-1 & top-5 & top-1 & top-5 & top-1 & top-5 \\
\midrule
B32
& 9.8 & 31.5 & 52.8 & 81.3 & 31.8 & 67.0 & 19.5 & 47.5 & 28.4 & 56.8 \\
RN50
 & 12.0 & 37.5 & 50.3 & 83.0 & 29.8 & 65.8 & 19.0 & 44.5 & 27.8 & 57.7 \\
VAE
& 2.9  & 13.2  & 32.5 & 65.3 & 12.4 & 39.1 & 5.6 & 19.1 & 13.4 & 34.2 \\
\midrule
RN50+B32
& 9.8 & 31.5 & 52.0 & 83.0 & 32.5 & 67.8 & 18.8 & 47.8 & 28.3 & 57.5 \\
RN50+VAE
& 9.0 & 22.8 & 48.0 & 85.3 & 26.5 & 61.3 & 11.5 & 30.3 & 23.8 & 49.9 \\
B32+VAE
& 12.0 & 33.5 & 64.5 & 91.3 & 39.0 & 76.8 & 17.3 & 43.8 & 33.2 & 61.3 \\
\midrule
RN50+B32+VAE
& 14.0 & 31.8 & 63.8 & 91.8 & 41.0 & 78.3 & 17.0 & 41.0 & 33.9 & 60.7 \\
\bottomrule
\end{tabular}}
\end{table*}

%% file: tabs/appendix_recon_eeg_h14.tex
\begin{table*}[t]
\centering
\caption{Reconstruction metrics across subjects using the H14 setting (higher $\uparrow$ is better, lower $\downarrow$ is better).}
\label{tab:eeg_recon_h14}
\footnotesize
\setlength{\tabcolsep}{5pt}
\renewcommand{\arraystretch}{1.15}
\begin{tabular}{c ccc cccc}
\toprule
& \multicolumn{2}{c}{\textbf{Low-level}} & \multicolumn{5}{c}{\textbf{High-level}} \\
\cmidrule(lr){2-3}\cmidrule(lr){4-8}
\textbf{Subject} 
& Pixcorr$\uparrow$ & SSIM$\uparrow$ & AlexNet(2)$\uparrow$ 
& AlexNet(5)$\uparrow$ & Inception$\uparrow$ & CLIP$\uparrow$ & SwAV$\downarrow$ \\
\midrule
1  & 0.179 & 0.305 & 0.828 & 0.870 & 0.719 & 0.732 & 0.588 \\
2  & 0.174 & 0.331 & 0.826 & 0.868 & 0.712 & 0.769 & 0.588 \\
3  & 0.177 & 0.317 & 0.832 & 0.872 & 0.703 & 0.802 & 0.574 \\
4  & 0.167 & 0.326 & 0.803 & 0.863 & 0.752 & 0.778 & 0.573 \\
5  & 0.163 & 0.315 & 0.804 & 0.846 & 0.676 & 0.745 & 0.593 \\
6  & 0.181 & 0.316 & 0.838 & 0.874 & 0.715 & 0.763 & 0.588 \\
7  & 0.155 & 0.328 & 0.811 & 0.874 & 0.736 & 0.775 & 0.569 \\
8  & 0.193 & 0.349 & 0.852 & 0.906 & 0.781 & 0.795 & 0.550 \\
9  & 0.163 & 0.330 & 0.820 & 0.872 & 0.765 & 0.762 & 0.561 \\
10  & 0.192 & 0.350 & 0.837 & 0.870 & 0.773 & 0.810 & 0.556 \\
\midrule
Ave & 0.174 & 0.327 & 0.825 & 0.871 & 0.733 & 0.773 & 0.574 \\
\bottomrule
\end{tabular}
\end{table*}

%% file: tabs/appendix_recon_eeg_h14+b32.tex
\begin{table*}[t]
\centering
\caption{Reconstruction metrics across subjects using the H14{+}B32 setting (higher $\uparrow$ is better, lower $\downarrow$ is better).}
\label{tab:eeg_recon_h14+b32}
\footnotesize
\setlength{\tabcolsep}{5pt}
\renewcommand{\arraystretch}{1.15}
\begin{tabular}{c ccc cccc}
\toprule
& \multicolumn{2}{c}{\textbf{Low-level}} & \multicolumn{5}{c}{\textbf{High-level}} \\
\cmidrule(lr){2-3}\cmidrule(lr){4-8}
\textbf{Subject} 
& Pixcorr$\uparrow$ & SSIM$\uparrow$ & AlexNet(2)$\uparrow$ 
& AlexNet(5)$\uparrow$ & Inception$\uparrow$ & CLIP$\uparrow$ & SwAV$\downarrow$ \\
\midrule
1  & 0.193 & 0.317 & 0.816 & 0.886 & 0.761 & 0.771 & 0.568 \\
2  & 0.190 & 0.346 & 0.845 & 0.919 & 0.789 & 0.821 & 0.551 \\
3  & 0.191 & 0.330 & 0.834 & 0.903 & 0.758 & 0.827 & 0.559 \\
4  & 0.183 & 0.334 & 0.825 & 0.905 & 0.805 & 0.840 & 0.535 \\
5  & 0.176 & 0.326 & 0.825 & 0.903 & 0.720 & 0.795 & 0.561 \\
6  & 0.191 & 0.326 & 0.833 & 0.907 & 0.791 & 0.817 & 0.552 \\
7  & 0.166 & 0.337 & 0.831 & 0.910 & 0.765 & 0.794 & 0.556 \\
8  & 0.207 & 0.365 & 0.861 & 0.918 & 0.815 & 0.827 & 0.528 \\
9  & 0.183 & 0.348 & 0.838 & 0.904 & 0.792 & 0.797 & 0.535 \\
10  & 0.190 & 0.365 & 0.848 & 0.921 & 0.830 & 0.854 & 0.521 \\
\midrule
Ave & 0.187 & 0.339 & 0.836 & 0.908 & 0.783 & 0.814 & 0.547 \\
\bottomrule
\end{tabular}
\end{table*}

%% file: tabs/appendix_recon_eeg_h14+vae.tex
\begin{table*}[t]
\centering
\caption{Reconstruction metrics across subjects using the H14{+}VAE setting (higher $\uparrow$ is better, lower $\downarrow$ is better).}
\label{tab:eeg_recon_h14+vae}
\footnotesize
\setlength{\tabcolsep}{5pt}
\renewcommand{\arraystretch}{1.15}
\begin{tabular}{c ccc cccc}
\toprule
& \multicolumn{2}{c}{\textbf{Low-level}} & \multicolumn{5}{c}{\textbf{High-level}} \\
\cmidrule(lr){2-3}\cmidrule(lr){4-8}
\textbf{Subject} 
& Pixcorr$\uparrow$ & SSIM$\uparrow$ & AlexNet(2)$\uparrow$ 
& AlexNet(5)$\uparrow$ & Inception$\uparrow$ & CLIP$\uparrow$ & SwAV$\downarrow$ \\
\midrule
1  & 0.156 & 0.301 & 0.755 & 0.762 & 0.653 & 0.646 & 0.658 \\
2  & 0.175 & 0.323 & 0.801 & 0.852 & 0.643 & 0.743 & 0.608 \\
3  & 0.171 & 0.290 & 0.793 & 0.853 & 0.651 & 0.730 & 0.621 \\
4  & 0.167 & 0.307 & 0.798 & 0.851 & 0.707 & 0.761 & 0.589 \\
5  & 0.174 & 0.295 & 0.783 & 0.847 & 0.670 & 0.723 & 0.611 \\
6  & 0.174 & 0.304 & 0.806 & 0.845 & 0.669 & 0.720 & 0.612 \\
7  & 0.154 & 0.315 & 0.764 & 0.828 & 0.655 & 0.711 & 0.622 \\
8  & 0.196 & 0.335 & 0.817 & 0.859 & 0.691 & 0.734 & 0.590 \\
9  & 0.166 & 0.315 & 0.765 & 0.827 & 0.678 & 0.701 & 0.605 \\
10  & 0.194 & 0.337 & 0.810 & 0.856 & 0.703 & 0.746 & 0.594 \\
\midrule
Ave & 0.173 & 0.312 & 0.789 & 0.838 & 0.672 & 0.721 & 0.611 \\
\bottomrule
\end{tabular}
\end{table*}

%% file: tabs/appendix_recon_eeg_h14+b32+vae.tex
\begin{table*}[t]
\centering
\caption{Reconstruction metrics across subjects using the H14{+}B32{+}VAE setting (higher $\uparrow$ is better, lower $\downarrow$ is better).}
\label{tab:eeg_recon_h14+b32+vae}
\footnotesize
\setlength{\tabcolsep}{5pt}
\renewcommand{\arraystretch}{1.15}
\begin{tabular}{c ccc cccc}
\toprule
& \multicolumn{2}{c}{\textbf{Low-level}} & \multicolumn{5}{c}{\textbf{High-level}} \\
\cmidrule(lr){2-3}\cmidrule(lr){4-8}
\textbf{Subject} 
& Pixcorr$\uparrow$ & SSIM$\uparrow$ & AlexNet(2)$\uparrow$ 
& AlexNet(5)$\uparrow$ & Inception$\uparrow$ & CLIP$\uparrow$ & SwAV$\downarrow$ \\
\midrule
1  & 0.193 & 0.332 & 0.835 & 0.883 & 0.727 & 0.757 & 0.578 \\
2  & 0.188 & 0.341 & 0.846 & 0.901 & 0.769 & 0.807 & 0.559 \\
3  & 0.196 & 0.324 & 0.834 & 0.900 & 0.755 & 0.826 & 0.566 \\
4  & 0.187 & 0.320 & 0.821 & 0.903 & 0.774 & 0.824 & 0.553 \\
5  & 0.179 & 0.317 & 0.831 & 0.893 & 0.705 & 0.797 & 0.565 \\
6  & 0.211 & 0.329 & 0.852 & 0.920 & 0.758 & 0.811 & 0.561 \\
7  & 0.179 & 0.336 & 0.833 & 0.914 & 0.754 & 0.805 & 0.554 \\
8  & 0.219 & 0.356 & 0.876 & 0.926 & 0.788 & 0.827 & 0.531 \\
9  & 0.198 & 0.342 & 0.842 & 0.889 & 0.757 & 0.783 & 0.546 \\
10  & 0.203 & 0.358 & 0.858 & 0.922 & 0.777 & 0.846 & 0.530 \\
\midrule
Ave & 0.195 & 0.336 & 0.843 & 0.905 & 0.756 & 0.808 & 0.554 \\
\bottomrule
\end{tabular}
\end{table*}

%% file: tabs/appendix_extended_visual.tex
\begin{table}[t]
  \centering
  \caption{
    % \textcolor{BurntOrange}{Top-1 and top-5 accuracy (\%) for 200-way zero-shot retrieval on THINGS-EEG under extended visual encoder settings. The first two columns report intra-subject performance, and the last two columns report inter-subject performance. Results are averaged over all 10 subjects under the same training configuration, with only the visual encoder stack varied. Numbers in parentheses denote absolute gains over the corresponding single-encoder baseline.}
    Top-1 and top-5 accuracy (\%) for 200-way zero-shot retrieval on THINGS-EEG under extended visual encoder settings. The first two columns report intra-subject performance, and the last two columns report inter-subject performance. Results are averaged over all 10 subjects under the same training configuration, with only the visual encoder stack varied. Numbers in parentheses denote absolute gains over the corresponding single-encoder baseline.
  }
  % \small
  \setlength{\tabcolsep}{6pt}
  \begin{tabular}{lcccc}
    \toprule
    & \multicolumn{2}{c}{Intra-subject} & \multicolumn{2}{c}{Inter-subject} \\
    \cmidrule(lr){2-3}
    \cmidrule(lr){4-5}
    Visual setting & top-1 & top-5 & top-1 & top-5 \\
    \midrule
    DINO               & 23.5 & 50.9 & 10.7  & 22.8  \\
    SynCLR             & 68.0 & 91.8 & 19.5  & 42.0  \\
    \midrule
    DINO+RN50           & 28.6 {\scriptsize (+5.1)} & 57.7 {\scriptsize (+6.8)} & 11.9  {\scriptsize (+1.2)} & 26.3  {\scriptsize (+3.5)} \\
    DINO+B32           & 35.7 {\scriptsize (+12.2)} & 66.8 {\scriptsize (+15.9)} & 13.3  {\scriptsize (+2.6)} & 29.9 {\scriptsize (+7.1)} \\
    DINO+VAE           & 50.1 {\scriptsize (+\textbf{26.6})} & 79.9 {\scriptsize (+\textbf{29.0})} & 16.1 {\scriptsize (+\textbf{5.4})} & 34.8 {\scriptsize (+\textbf{12.0})} \\
    SynCLR+RN50         & 69.1  {\scriptsize (+1.1)} & 91.9  {\scriptsize (+0.1)} & 19.2 {\scriptsize (-0.3)} & 41.0  {\scriptsize (-1.0)} \\
    SynCLR+B32         & 68.4  {\scriptsize (+0.4)} & 92.4 {\scriptsize (+0.6)} & 19.2  {\scriptsize (-0.3)} & 42.3  {\scriptsize (+0.3)} \\
    SynCLR+VAE         & 78.5 {\scriptsize (+\textbf{10.5})} & 96.4 {\scriptsize (+\textbf{4.6})} & 23.0 {\scriptsize (+\textbf{3.5})} & 46.5 {\scriptsize (+\textbf{4.5})} \\
    \midrule
    DINO+RN50+VAE      & 53.3 {\scriptsize (+29.8)} & 82.6 {\scriptsize (+31.7)} & 17.3 {\scriptsize (+6.6)}  & 35.8 {\scriptsize (+13.0)} \\
    DINO+B32+VAE       & 58.5 {\scriptsize (+\textbf{35.0})} & 85.9 {\scriptsize (+\textbf{35.0})} & 19.3 {\scriptsize (+\textbf{8.6})} & 39.8 {\scriptsize (+\textbf{17.0})} \\
    SynCLR+RN50+VAE    & 78.7 {\scriptsize (+10.7)} & 96.5 {\scriptsize (+4.7)} & 23.4 {\scriptsize (+3.9)} & 47.1 {\scriptsize (+5.1)} \\
    SynCLR+B32+VAE     & 78.9 {\scriptsize (+\textbf{10.9})} & 96.5 {\scriptsize (+\textbf{4.7})} & 23.4  {\scriptsize (+\textbf{3.9})}& 47.2 {\scriptsize (+\textbf{5.2})} \\
    \bottomrule
  \end{tabular}
  \label{tab:visual_ablation_extended}
\end{table}

%% file: tabs/appendix_channels.tex
\begin{table*}[t]
  \centering
  \small
  \setlength{\tabcolsep}{3pt}
  \caption{
    % \textcolor{BurntOrange}{Top-1 and top-5 accuracy (\%) for 200-way zero-shot retrieval on THINGS-EEG across visual-encoder settings (rows) and channel selections (columns). Numbers are averaged over all 10 subjects under the same training configuration, with only the selected channels varied.}
    Top-1 and top-5 accuracy (\%) for 200-way zero-shot retrieval on THINGS-EEG across visual-encoder settings (rows) and channel selections (columns). Numbers are averaged over all 10 subjects under the same training configuration, with only the selected channels varied.
  }
  \resizebox{\textwidth}{!}{%
  \begin{tabular}{l*{5}{cc}}
    \toprule
    & \multicolumn{2}{c}{Occipital}
    & \multicolumn{2}{c}{Parietal}
    & \multicolumn{2}{c}{O+P (Our)}
    & \multicolumn{2}{c}{Others}
    & \multicolumn{2}{c}{All} \\
    \cmidrule(lr){2-3}
    \cmidrule(lr){4-5}
    \cmidrule(lr){6-7}
    \cmidrule(lr){8-9}
    \cmidrule(lr){10-11}
    Method
    & top-1 & top-5
    & top-1 & top-5
    & top-1 & top-5
    & top-1 & top-5
    & top-1 & top-5 \\
    \midrule
    RN50         & 43.1 & 75.51 & 21.8  & 52.1  & 48.1  & 80.5  &  9.4 &	26.9   & 43.1 & 75.7  \\
    B32          & 48.3 & 80.0  & 23.9  & 52.8  & 51.2  & 83.0  &  8.5 &	26.4   & 44.1  & 78.3 \\
    VAE          &45.9  & 76.6  & 10.0  & 27.7  & 44.1  & 75.5  &  2.8 &	10.9   & 27.0  & 59.0  \\
    \midrule
    RN50+B32     & 51.9   & 83.2  & 25.6  & 54.7  & 56.2  & 86.2  &  8.7 &	27.2   &  47.4 	& 80.8  \\
    RN50+VAE     & 65.1  & 90.3  & 19.8  & 47.1  & 65.2  & 90.7  &  5.9 	&18.4   &  43.6 	&76.8   \\
    B32+VAE   & 71.3  & 93.1 & 26.2  & 57.1  & 73.7  & 94.1  &  9.3 & 27.2   &  59.7 &	88.9  \\
    \midrule
    RN50+B32+VAE & 72.6  &  93.5  & 27.4  & 58.9  & 75.7  & 94.6  &  9.1 &	28.4   &  62.5 	&90.0  \\
    \bottomrule
  \end{tabular}%
  }

  \label{tab:channel_ablation}
\end{table*}

%% file: tabs/appendix_mask.tex
\begin{table}[t]
\centering
\caption{
% \textcolor{BurntOrange}{Encoder-masking ablation for brain-to-image retrieval with the RN50+B32+VAE fusion setting. We report average top-1/top-5 accuracy (\%) for 200-way zero-shot retrieval on THINGS-EEG, masking one visual encoder at inference under identical training configurations.}
Encoder-masking ablation for brain-to-image retrieval with the RN50+B32+VAE fusion setting. We report average top-1/top-5 accuracy (\%) for 200-way zero-shot retrieval on THINGS-EEG, masking one visual encoder at inference under identical training configurations.
}
\label{tab:encoder_masking}
\footnotesize
\setlength{\tabcolsep}{6pt}
\renewcommand{\arraystretch}{1.05}
\begin{tabular}{lcccc}
\toprule
\multirow{2}{*}{Setting} 
& \multicolumn{2}{c}{Intra-subject} 
& \multicolumn{2}{c}{Inter-subject} \\
& Top-1 & Top-5 & Top-1 & Top-5 \\
\midrule
w/o RN50 & 70.6 & 92.7 & 19.8 & 41.2 \\
w/o B32  & 41.2 & 73.2 & 11.9 & 27.4 \\
w/o VAE  & 29.6 & 60.2 &  9.3 & 23.4 \\
FULL     & \textbf{75.7} & \textbf{94.6} & \textbf{20.0} & \textbf{44.1} \\
\bottomrule
\end{tabular}
\end{table}